\documentclass[10pt,journal,compsoc]{IEEEtran}
\ifCLASSOPTIONcompsoc
  \usepackage[nocompress]{cite}
\else
  \usepackage{cite}
\fi
\ifCLASSINFOpdf
  \usepackage[pdftex]{graphicx}
\else
\fi
\usepackage{amsmath}
\usepackage{algorithm}
\usepackage{algorithmic}

\usepackage{verbatim}

\usepackage{url}

\usepackage{hyperref}       
\usepackage{url}            
\usepackage{booktabs}       
\usepackage{amsfonts}       
\usepackage{microtype}      
\usepackage{xcolor}         
\usepackage{xspace}
\usepackage{bbold}
\usepackage{multirow}
\usepackage{bbding}
\usepackage{pifont}
\usepackage{algorithm}  
\usepackage{algorithmic}  

\makeatletter
\DeclareRobustCommand\onedot{\futurelet\@let@token\@onedot}
\def\@onedot{\ifx\@let@token.\else.\null\fi\xspace}
\def\eg {\emph{e.g}\onedot} 
\def\ie{\emph{i.e}\onedot} 
 
\def\etc{\emph{etc}\onedot} 
 
\def\etal{\emph{et al}\onedot}
\makeatother
\usepackage{balance}

\begin{document}
%
\title{Safety and Performance, Why Not Both? Bi-Objective Optimized Model Compression against Heterogeneous Attacks \\ Toward AI Software Deployment}
%
%
%
%

\author{Jie~Zhu,
        Leye~Wang, 
        Xiao~Han, 
        Anmin~Liu,
        and~Tao~Xie,~\IEEEmembership{Fellow,~IEEE}
\IEEEcompsocitemizethanks{\IEEEcompsocthanksitem Jie~Zhu, Leye~Wang, Anmin~Liu, and~Tao~Xie are with Key Lab of High Confidence Software Technologies (Peking University), Ministry of Education, China, and School of Computer Science, Peking University, China.\protect\\
E-mail: zhujie@stu.pku.edu.cn, leyewang@pku.edu.cn, anminliu@stu.pku.edu.cn, taoxie@pku.edu.cn
\IEEEcompsocthanksitem Xiao Han is with Shanghai University of Finance and Economics, China. \protect\\ 
E-mail: xiaohan@shufe.edu.cn \protect\\
 Corresponding author: Leye Wang and Xiao Han.
}

\thanks{Manuscript received April 19, 2005; revised August 26, 2015.}}

%
%

\markboth{Journal of \LaTeX\ Class Files,~Vol.~14, No.~8, August~2015}%
{Shell \MakeLowercase{\textit{et al.}}: Bare Demo of IEEEtran.cls for Computer Society Journals}
%



\IEEEtitleabstractindextext{%
\begin{abstract} 
The size of deep learning models in artificial intelligence (AI) software is increasing rapidly, hindering the large-scale deployment on resource-restricted devices (\eg, smartphones).
To mitigate this issue, AI software compression plays a crucial role, which aims to compress model size while keeping high performance. However, the intrinsic defects in a  big model may be inherited by the compressed one. Such defects may be easily leveraged by adversaries, since a compressed model is usually deployed in a large number of devices without adequate protection. In this article, we aim to address the safe model compression problem from the perspective of safety-performance co-optimization. Specifically, inspired by the test-driven development (TDD) paradigm in software engineering, we propose a test-driven sparse training framework called \textit{SafeCompress}. By simulating the attack mechanism as safety testing, SafeCompress can automatically compress a big model to a small one following the dynamic sparse training paradigm. 
Then, considering two kinds of representative and heterogeneous attack
mechanisms, \ie, black-box membership inference attack and white-box membership inference attack, we develop two concrete instances called BMIA-SafeCompress and WMIA-SafeCompress. Further, we implement another instance called MMIA-SafeCompress by
extending SafeCompress to defend against the occasion when adversaries conduct black-box and white-box membership inference attacks simultaneously. We conduct extensive experiments on five datasets for both computer vision and natural language processing tasks. The results show  the effectiveness and generalizability of our framework. We also discuss how to adapt SafeCompress to other attacks besides membership inference attack, demonstrating the flexibility of SafeCompress.
\end{abstract}

}

\maketitle
\IEEEdisplaynontitleabstractindextext

%
\IEEEpeerreviewmaketitle

\IEEEraisesectionheading{\section{Introduction}\label{sec:introduction}}

%
%
%
%
\IEEEPARstart{I}{n} the last decade, artificial intelligence (AI) software, especially software based on deep neural networks (DNN), has attracted much attention and made a significant influence~\cite{alexnet_krizhevsky2012imagenet}. Currently, AI software, with DNN as representatives, is recognized as an emerging type of software artifact (sometimes known as ``software 2.0”~\cite{remos_zhang2022remos}). Notably, the size of DNN-based AI software has increased rapidly in recent years (mostly because of a trained deep neural network model). For instance, a state-of-the-art model of computer vision  contains more than 15 billion parameters~\cite{riquelme2021scaling}. A recent natural language model, GPT-3, is even bigger, surpassing 175 billion parameters; this situation requires nearly 1TB of space to store only the model~\cite{brown2020language}. Such a big model hinders realistic applications such as autonomous driving when the software is required to be deployed in resource-restricted devices such as wearable devices or edge nodes. 
To this end, a new branch is derived from the traditional area of software compression~\cite{Pugh1999CompressingJC,drinic2007ppmexe}, called \textit{AI software compression} (especially DNN \textit{model compression}\footnote{In the rest of the article, without incurring ambiguity, we use `model compression' to indicate `DNN model compression in AI software' for brevity.}), and has attracted a lot of research interest.

Model compression aims to compress a big DNN model to a smaller one given specific requirements, \eg, parameter numbers, model sparsity, and compression rate. Rashly compressing a model may lead to severe degeneration in the AI software's task performance such as classification accuracy. To balance memory storage and task performance, many compression approaches have been proposed and deployed~\cite{Tiny_BERT_jiao2019tinybert, quantify_han2015learning}. 
For example, Han~\etal~\cite{quantify_han2015learning} prune AlexNet~\cite{alexnet_krizhevsky2012imagenet} and reduce its size  by 9 times while losing only 0.01\% accuracy in image classification. Jiao~\etal~\cite{Tiny_BERT_jiao2019tinybert} reduce the size of BERT~\cite{devlin2018bert} by about 2 times via knowledge distillation while losing 0.10\%  average score in the GLUE~\cite{wang2018glue} official benchmark.

While a compressed model aims to mimic the original model's behavior, its defects may also be inherited. As a representative case, big deep models are verified to be able to memorize training data~\cite{secret_carlini2019secret}, thus leading to private data leakage when facing threats such as membership inference attack~\cite{Shaow_Learning}; such a vulnerability would probably remain in the compressed model. More seriously, model compression is often used for AI software deployment on a large number of edge devices (smartphones, wearable devices, \etc)~\cite{Deng2020ModelCA, liu2020ensemble}; compared to big models (often stored in a well-maintained server), an adversary\footnote{Without incurring ambiguity, we use `adversary' to indicate a person who conducts the attack and use `attacker' to indicate the attack model an adversary utilizes.} thereby has much more opportunities to gain access to compressed models and attack them. For example, an adversary may act as a normal user to download a compressed model on her/his own device. A typical scenario is  using a virtual shopping assistant in a smartphone. If software developers neglect the privacy risk brought by the AI model in the virtual shopping assistant and directly deploy the assistant, when an adversary obtains a victim's smartphone, the adversary could attack the AI model in a certain way and analyze the output to acquire the victim's purchase preference and financial status. \textit{In a word, a compressed model inherits the vulnerabilities of the original big model, while facing even higher risks of being attacked\footnote{On one hand, accessing a compressed model is easier for adversaries. On the other hand, in our experiments, we observe that some compressed models, \eg, ones compressed by knowledge distillation (KD-AdvReg, Sparsity=0.2) in Figure~\ref{overview}, suffer higher attack accuracy (75.54\%) than uncompressed ones (67.33\%), indicating that compression could potentially result in a more vulnerable model.}}. Hence, studying how to conduct \textbf{safe model compression} is urgently required.

An intuitive solution to safe model compression is directly combining two streams of techniques, forming a \textit{two-step solution}: (i) model compression and (ii) model protection. For instance, we can first obtain a small model using existing compression techniques, and then apply protection techniques (\eg, differential privacy~\cite{dp_abadi2016} and knowledge distillation~\cite{kd_membership}) to improve the model safety against certain attacks. \textit{However, the two-step solution may suffer from poor model performance\footnote{We use `performance' as a general term to describe any specific task-dependent metric. For instance, if a model is developed for image classification, the model performance can be measured by the classification accuracy metric.} and low safety because protection techniques could fail to consider the potential variability introduced by model compression.} For example, Yuan and  Zhang~\cite{pruning_defeat_yuan2022membership} find that pruning makes the divergence of prediction confidence and sensitivity increase and vary widely among different classes under prediction. This phenomenon may not be sensed by defenders due to lack of interaction but can be manipulated by adversaries.

In this article, we aim to address the problem of safe model compression from the perspective of \textit{performance-safety co-optimization}. Specifically, inspired by \textit{test-driven development}~\cite{Beck2003TestdrivenD} and \textit{dynamic sparse training}~\cite{DST_Begin_mocanu2018scalable}, we propose a framework of \textit{test-driven sparse training} for \textit{safe model compression}, called \textbf{\textit{SafeCompress}}. By simulating the attack mechanism to defend, SafeCompress can automatically compress a big model into a small one (with required sparsity) to optimize both model performance and safety. SafeCompress generally follows an iterative optimization mechanism. For initialization, SafeCompress randomly prunes a big model to a sparse one, which serves as the input of the first iteration. In each iteration, SafeCompress applies various compression strategies to the input model to derive more sparse models. Then SafeCompress launches a performance-safety co-optimization mechanism that applies both task-performance evaluation and simulated-attack-based safety testing on the derived sparse models to select the best compression strategy. Then, the sparse model with the selected (best) strategy becomes the input of the next iteration. The iterative process will terminate after a predefined maximum number of iterations or a new model has little improvement in performance evaluation and safety testing. 

Utilizing the SafeCompress framework, our work addresses the challenge of safeguarding compressed AI models in software against heterogeneous attacks that frequently occur in reality. Specifically, in this work, we first consider a single attack and then extend to multiple heterogeneous attacks. In the case of a single attack, we specifically consider two representative privacy attacks, namely, black-box~\cite{Shaow_Learning} and white-box membership inference attacks (MIAs)~\cite{sablayrolles2019white}, and implement two concrete safe model compression mechanisms against them, denoted as \textit{\textbf{BMIA-SafeCompress}} and \textit{\textbf{WMIA-SafeCompress}}, respectively, which effectively mitigate the risks associated with the respective MIA. 
Note that we choose MIA as the attack example because MIA is representative of evaluating AI model safety, especially from the aspect of privacy leakage~\cite{dp_yeom2018privacy, nasr2019comprehensive}. For multiple heterogeneous attacks, our experiments (Section~\ref{sec:experiemnts}) show  that our SafeCompress framework is highly flexible and generalizable, naturally supporting extensions to defend against multiple heterogeneous attacks. Specifically, we incorporate additional attack simulation techniques to SafeCompress and implement a concrete instance called \textit{\textbf{MMIA-SafeCompress}} to defend against \textit{heterogeneous} black-box and white-box MIAs\footnote{Heterogeneous black-box and white-box MIAs are those that construct attack models with different structures~\cite{nasr2019comprehensive, MIAs_Survey2021} (as illustrated in Figure~\ref{attack}) so as to operate under varying degrees of knowledge about a target model.
Specifically, the black-box setting posits that only model outputs are available to the adversary while in the white-box setting, besides model outputs, extra knowledge about the target model (\eg, hidden layer parameters~\cite{sablayrolles2019white} and gradient descents in training epochs~\cite{nasr2019comprehensive}) is also obtainable.} concurrently. Furthermore, we integrate adversarial training into SafeCompress by including attack models in the training process to enhance a compressed model's defense capability.
	
This article makes the following main contribution:

$\bullet$ To the best of our knowledge, this work is the first effort toward the problem of \textbf{safe model compression}, which is critical for today's large-scale AI software deployment on edge devices such as smartphones.

$\bullet$ To address the problem of safe model compression, we propose a general framework called SafeCompress, which can be configured to protect against a pre-specified attack mechanism. SafeCompress adopts a test-driven process to iteratively update model compression strategies to co-optimize model performance and safety. 

$\bullet$ We take into account a single attack and multiple heterogeneous attacks, and choose representative MIAs as attack mechanisms~\cite{ResAdv}. For a single attack, we develop two concrete instances of SafeCompress, \ie, BMIA-SafeCompress and WMIA-SafeCompress. For multiple heterogeneous attacks, we extend SafeCompress and implement MMIA-SafeCompress. Additionally, we integrate adversarial training into SafeCompress to further enhance a compressed  model’s
defense capability. We also discuss how to adapt SafeCompress to other attacks.

$\bullet$ Using BMIA-SafeCompress, WMIA-SafeCompress, and MMIA-SafeCompress as showcases of SafeCompress, we conduct extensive experiments on five datasets of two domains (three tasks of computer vision and two tasks of natural language processing). The experimental results show that our framework\footnote{Without incurring ambiguity, we use `framework' to indicate `general framework' and `approach' to indicate `constructed instance based on SafeCompress or other specific approaches'.} significantly outperforms baseline solutions that integrate state-of-the-art compression and MIA defense approaches. The code of the three SafeCompress instances is available as open source at \url{https://github.com/JiePKU/SafeCompress}.

This article is an extension of a previous conference edition~\cite{zhu2022safety}. Compared to the conference edition, this article includes six main improvements: (1) configuring Safecompress to defend against white-box membership inference attack, (2) extending SafeCompress to handle multiple heterogeneous attacks, (3) incorporating
adversarial training into SafeCompress, (4) performing robustness analyses on the safety and performance trade-off metric, (5) adding a new baseline called KL-Div, and (6) involving more recent work discussed in the related work. 

\section{Background}
In this section, we briefly introduce contextual knowledge related to our work

\subsection{Test-Driven Development}
Test-driven development (TDD)~\cite{Beck2003TestdrivenD} is a programming paradigm where test code plays a vital role during the whole software development process. With TDD, before writing code for  software functionality, programmers could write the corresponding test suite in advance; then, the test suite can justify whether the software functionality is implemented properly or not. In general, TDD leads to an \textit{iterative coding-testing} process to improve the correctness and robustness of the software; TDD has become a widely adopted practice in software development (\eg, SciPy~\cite{Virtanen2020SciPy1F}). Inspired by TDD, we develop a test-driven framework of safe model compression called SafeCompress. Similar to the iterative coding-testing process in TDD, SafeCompress adopts an \textit{iterative compressing-testing} process. In particular, by specifying the attacks to fight against, SafeCompress can automatically update the compression strategies step by step to optimize model performance and safety simultaneously.

\subsection{Dynamic Sparse Training}
Dynamic sparse training (DST) is a sparse-to-sparse training paradigm to learn a sparse (small) DNN model based on a dense (big) one~\cite{DST_Begin_mocanu2018scalable, mocanu2017network}. Specifically, DST starts training with a sparse model structure initialized from a dense model.  As training progresses, it modifies the architecture iteratively by pruning some neural network connections and growing new connections based on certain strategies. This manner enables neural networks to explore self-structure until finding the most suitable one for training data. SafeCompress's iterative updating strategy for optimizing a compressed model structure is inspired by the DST paradigm. 

\subsection{Membership Inference Attack}
Prior research has extensively shown 
that DNN models often exhibit different behaviors on training data records (\ie, members) versus test data records (\ie, non-members); the main reason is that models can memorize training data during the repeated training process lasting for a large number of epochs~\cite{secret_carlini2019secret,nasr2019comprehensive}. For instance, a DNN model can  generally give a higher prediction confidence score to a training data record than a test one, as the model may remember the training data's labels. Based on such observations, membership inference attack (MIA)~\cite{Shaow_Learning} is proposed to build attack model to infer whether one data record belongs to training data or not. When it comes to sensitive data,  personal privacy is exposed to a great risk. For example, if MIA learns that a target user's electronic health record (EHR) data is used to train a model related to a specific disease (\eg, to predict the length of stay in ICU~\cite{ma2021distilling}), then the adversary knows that the target user has the disease. In this work, we use MIA to empirically demonstrate  the feasibility and effectiveness of SafeCompress, because MIA has become one of the representative attacks to evaluate the safety of DNN models both theoretically and empirically~\cite{dp_yeom2018privacy, nasr2019comprehensive}. Moreover, compressed models in edge devices are easily accessible, \eg, by downloading, allowing adversaries to snoop and collect more model information. Hence, when compressed models are deployed in edge devices, MIA could inflict great harm to personal privacy.

\subsection{Adversarial Training}

 Adversarial training~\cite{goodfellow2020generative, ResAdv} is a competitive learning approach that involves a target model and adversarial models with the aim of equipping the target model with the ability to withstand adversarial attacks. To achieve this ability, this training approach typically incorporates weighted objective terms of adversarial models into the loss function of the target model. Throughout the training process, the goal is to optimize the objective terms of the adversarial models while simultaneously minimizing the overall loss including both the target and adversarial objectives.  More generally, this approach is essentially a min-max game~\cite{madry2017towards} that has been widely used in computer vision~\cite{goodfellow2020generative, zhu2017unpaired} and natural language processing~\cite{miyato2016adversarial, shafahi2019adversarial}. In this work, this useful skill is combined with our SafeCompress framework to further help improve a compressed model's defense ability.  

\section{Problem Formulation}

Given a big model $\mathcal{F}(;\theta)$ parameterized by $\theta$, we aim to find a sparse model $\mathcal{F}(;\hat \theta)$ (most elements in $\hat \theta$ are zero) under certain memory restriction $\Omega$ and the sparse one can defend against a pre-specified attack mechanism $f_A$. We use $G_{f_A}$ to denote the attack gain of $f_A$. We restrict the compression ratio, also named as model sparsity, below $\Omega$ (\ie, the percentage of non-zero parameters in the sparse model over the original model).
We aim to \textit{minimize} both the task performance loss $\mathcal L$ and the attack gain $G_{f_A}$ over the sparse model $\mathcal{F}(;\hat \theta)$:
\begin{align} 
\label{general_problem_L}
& \min_{\hat \theta} \sum_ {x,y}  \mathcal{L} (\mathcal{F}(x;\hat \theta), y) \\
\label{general_problem_G}
& \min_{\hat \theta} G_{f_A} (\mathcal{F}(;\hat \theta)) \\
\label{general_problem_S}
& \operatorname{ s.t. }  \frac{{\; \lVert \hat \theta \rVert}_{0}}{{\lVert \theta \rVert}} \leq \Omega,
\end{align}
where $x$ is a sample and $y$ is the corresponding label, and $\mathcal{L}$ represents a
task-dependent loss function. ${\lVert \cdot  \lVert}_{0}$ counts the number of non-zero elements, and ${\lVert \cdot  \lVert}$ calculates the number of all the elements. This formulation is a general one without specifying the attack mechanism $f_A$ and is a bi-objective optimization problem regarding both model performance (Eq.~\ref{general_problem_L}) and safety (Eq.~\ref{general_problem_G}) when compressing big models.  

\textbf{Threat Model.} We take MIA, being the representative privacy attack, as a threat model to evaluate a victim model (\ie, a compressed model) about its leaked privacy. MIA usually trains a binary classification model $f_A$. In this work, we use a neural network classifier as $f_A$.

\textbf{MIA Gain}. The gain $G_{f_A}$ for MIA is then formulated as follows.  
Given training samples (${D}_{tr}$) and non-training samples (${D}_{\
	\neg tr}$), the expected gain of the threat model $f_A$ is 
\begin{equation} \label{G_}
\begin{split}
G_{f_A}(\mathcal F(;\hat \theta))= & \sum_ {x, y}  (\mathbb{1}(x \in D_{tr})\log (f_A(\mathcal F(x;\hat \theta),y)) +  \\
& \mathbb{1}(x \in D_{\neg tr})\log (1-(f_A(\mathcal F(x;\hat \theta),y)))) ,
\end{split}
\end{equation}
where $\mathbb{1}(x \in Z)$ is 1 if sample $x$ belongs to $Z$;  otherwise 0. 

\begin{figure*}[t]
	\centering{\includegraphics[width=.9\linewidth]{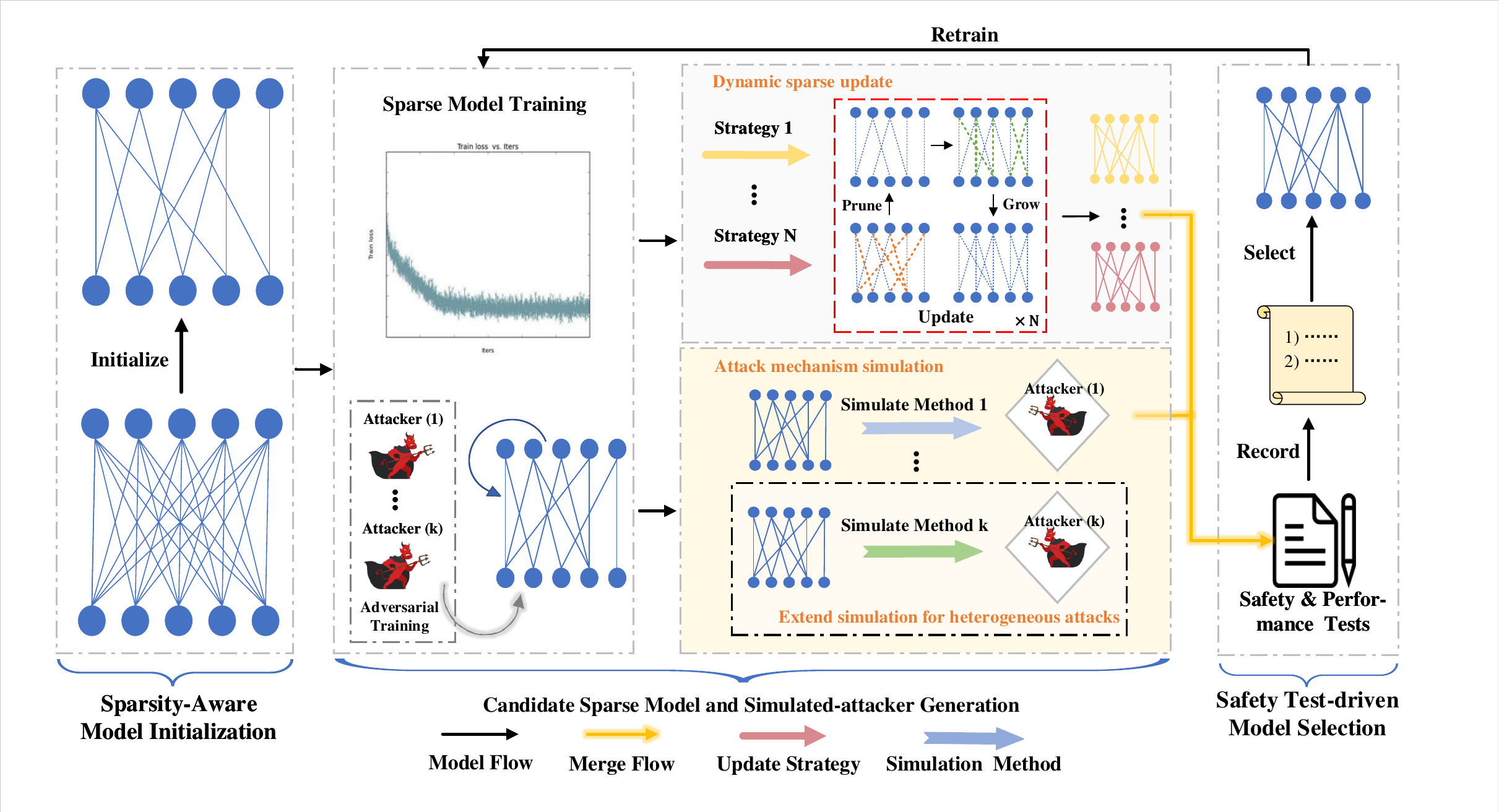}}
	\vspace{-0.4cm}
	\caption{An overview of our SafeCompress framework.}
	\label{overview}
	\vspace{-0.4cm}
\end{figure*}

\section{Approach}
In this section, we first describe our design principles. Then we present our SafeCompress framework. Finally, we show how to use our framework to defend against a single attack and  multiple heterogeneous attacks, respectively.
\subsection{Key Design Principles}
We illustrate two key principles driving our design, \ie, \textit{attack configurability} and \textit{task adaptability} before elaborating on design details.

\textbf{Attack Configurability}. In reality, adversaries may conduct various types of attacks~\cite{han2019f}. Hence, a practical solution should be able to do safe compression against an arbitrary pre-specified attack mechanism. In other words, the proposed solution should be easily configured to fight against a given attack mechanism.

\textbf{Task Adaptability}. AI (especially deep learning) techniques have been applied to various task domains including computer vision (CV), natural language processing (NLP), \etc. To this end, a useful solution is desired to be able to adapt to heterogeneous AI tasks (\eg, CV or NLP) very easily.

Our design of SafeCompress just follows these two principles, thus ensuring practicality in various scenarios of AI software deployment. Next, we describe the design details.

\subsection{SafeCompress: A General Framework for Safe Model Compression} \label{defense}
Figure~\ref{overview} presents an overview of SafeCompress, which contains three stages. We describe each below.

\textbf{Stage 1. Sparsity-Aware Model Initialization.} $\:$ During this stage, we follow the paradigm of dynamic sparse training to prioritize meeting the goal of sparsity (\ie, compression ratio).  
Specifically, based on a given big model (an arbitrary deep model for various tasks such as  CV or NLP), we initialize a sparser version of the model to align with specific memory requirements. After  initialization, the sparse model is sent to Stage 2.

\textbf{Stage 2. Candidate Sparse Model and Simulated-attacker Generation.} $\;$ During this stage, the sparse model is first trained until reaching the stopping criteria. Then, the trained model is fed to two branches. The first branch is called \textit{dynamic sparse update} where different combinations of pruning and growth strategies are performed on the input model, producing new candidate model variants with diverse sparse architectures. To ensure that the derived model always meets the given sparsity requirement, we keep the number of removed connections and reactivated ones the same in each update. The second branch is called \textit{attack mechanism simulation}. In this branch, we simulate an external attacker that aims to attack candidate sparse models. The simulated attacker and candidate sparse models are sent to Stage 3.  

\textbf{Stage 3. Safety Test-driven Model Selection.} $\;$ A safety testing is performed on these input sparse models using the simulated attacker. The candidate model that performs the best in this testing process will be selected and sent back to Stage 2, starting a new iteration.  
The whole process will terminate after running for a predefined number of iterations.
\begin{algorithm}[h]  
	\small
	\caption{SafeCompress Framework Procedure}  
	\label{alg:Framwork}  
	\begin{algorithmic}[1]  
		\REQUIRE A big model $\mathcal{M}_{L}$; a  sparsity requirement $\Omega$; a set of update strategies $\mathcal{U}$ whose  size is $N$; training stopping criteria for a sparse model $\mathcal{T}$; total epochs for termination $Eps$; 
		\ENSURE  
		A well-trained sparse model $\mathcal{M}_{S}$;
		\STATE Initialize $\mathcal{M}_{L}$ as a sparse one to meet the sparsity requirement $\Omega$; we denote this sparse model as $\mathcal{M}_{S}$
		\STATE Train $\mathcal{M}_{S}$ until  condition $\mathcal{T}$ is satisfied \textbf{then do}: \label{code:fram:extract}  
		\STATE \textbf{for} each $\mathcal{U}_{i}$ in $\mathcal{U}$ \textbf{do}:
		\STATE \quad Update $\mathcal{M}_{S}$ via $\mathcal{U}_{i}$ denoted as $\mathcal{M}_{S}^{i}$ 
		\STATE  \textbf{end for}	
		\STATE  Obtain a candidate sparse model set $\mathcal{C}$=\{$\mathcal{M}_{S}^{1}$...$\mathcal{M}_{S}^{N}$\}
		\STATE  Simulate an external attacker as $\mathcal{A}$ (with the help of $\mathcal{M}_{S}$)
		
		\STATE		$\mathcal{M}_{S}^{best}$ $\leftarrow$ Pick the best from \{safety testing($\mathcal{C}$, $\mathcal{A}$)\}
		\STATE $\mathcal{M}_{S}$ $\leftarrow$	$\mathcal{M}_{S}^{best}$
		\STATE \textbf{if not} achieve total epochs $Eps$:  	   			  
		\STATE \quad  \textbf{go to} \ref{code:fram:extract}
		\quad \RETURN $\mathcal{M}_{S}$  
	\end{algorithmic}  
\end{algorithm} 

The pseudo-code of SafeCompress is in Algorithm~\ref{alg:Framwork}. Note that we do not restrict the type of attacks in SafeCompress. It thus has the potential to prevent various attacks toward AI software and models. 

\subsection{Defense against Single Attack}
We consider scenarios where defenses are mounted against black-box MIA and white-box MIA, respectively.
\subsubsection{BMIA-SafeCompress: Defending Black-box Membership Inference Attack based on SafeCompress} \label{implementation}
We implement a concrete mechanism of safe model compression against black-box MIA\footnote{The black-box setting assumes that only model outputs are available to an adversary.}, called \textbf{\textit{BMIA-SafeCompress}} based on SafeCompress.

\textbf{Stage 1. Sparsity-Aware Model Initialization.} $\;$ Given a big model,  we adopt the Erdös–Rényi~\cite{initialization_method} initialization approach to reach a predefined sparsity requirement, \ie, removing a number of model connections (assigning zeros to the connection weights). Specifically, for the $k$-th layer with $n_k$ neurons, we collect them in a vector and denote it as $V^k = [v^{k}_{1}, v^{k}_{2}, v^{k}_{3},...... v^{k}_{n_k}]$. Usually, $V^k$ in the $k$-th layer is connected with the previous  layer $V^{k-1}$ via a weight matrix $W^k \in \mathbb{R}^{n_k \times n_{k-1}}$. In the sparse setting, the matrix degenerates to a Erdös–Rényi random graph, where the probability of connection between neuron $V^{k}$ and $V^{k-1}$ is decided by 
\begin{equation}
P(W_{ij}) = \frac{\epsilon \times (n_{k} + n_{k-1} )}{n_{k} \times n_{k-1}} \,,
\end{equation} 
where $\epsilon$ is a coefficient that is adjusted to meet a target sparsity.
In general, this distribution is inclined to allocate higher sparsity (more zeros) to layers with more parameters as the probability tends to decline while the quantity of parameters increases\footnote{For the weights of remaining  connections, we can keep those of the big model, or simply do random initialization. In our experiments, we find that keeping big model weights performs not better than random initialization. Hence, we adopt random initialization in our implementation.}.

\textbf{Stage 2. Candidate Sparse Model and Simulated-attacker Generation.} $\;$ During this stage, we train the input sparse model for a predefined number of iterations (following the setting in previous work~\cite{DST_InTime}, the number of iterations is set to 4,000). We can use a vanilla training strategy or adopt advanced adversarial training (we introduce it later in the end of this subsection). Once finished, the well-trained sparse model, denoted as $\mathcal{M}_{S}$, is fed into the two branches. 

\textit{Branch 1. Dynamic sparse update}. In the first branch, \ie, the branch of dynamic sparse update, we apply two state-of-the-art pruning strategies and two growth strategies to operate on $\mathcal{M}_{S}$, leading to four ($2*2$) different sparse typologies. The two pruning strategies are \textit{magnitude-based} pruning~\cite{pruning_han2015deep} and \textit{threshold-based} pruning~\cite{quantify_han2015learning}. Magnitude-based pruning removes connections with the smallest weight magnitudes (weight values); threshold-based pruning removes connections whose weight magnitudes are below a given threshold. The two pruning strategies are effective as small weight magnitudes often contribute little to the final output. The two growing strategies are \textit{gradient-based} growth~\cite{evci2020rigging} and \textit{random-based} growth~\cite{DST_Begin_mocanu2018scalable}. The gradient-based growth reactivates a connection (weight) that has a large gradient $\lvert \frac{\partial \mathcal{L}}{\partial w}\rvert$, which indicates that the weight is extremely eager to be updated.
The random-based growth randomly reactivates connections. This manner may help prevent the sparse model from getting stuck in a local optimum. Afterward, we fine-tune these derived sparse models and generate four candidate sparse models $\mathcal M_{S}^1,...,\mathcal M_{S}^4$\footnote{As more advanced pruning or growth strategies may be proposed in the future, SafeCompress is easy to incorporate them by adding to/replacing existing strategies.}.

\begin{figure} 
		\centering{\includegraphics[width=1\linewidth]{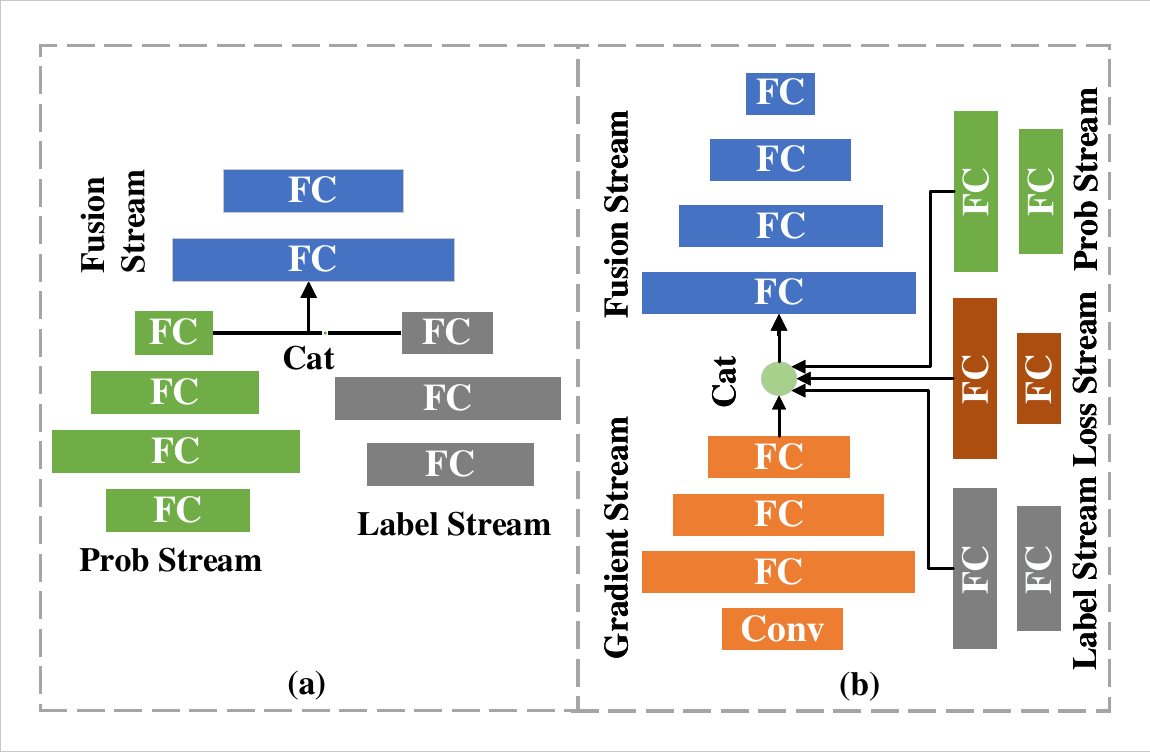}}
		\vspace{-0.7cm}
		\caption{The architecture of the simulated black-box MIA network model (a) and white-box one (b). FC is a fully connection layer. Conv is a convolution layer.}
		\label{attack}
		\vspace{-0.5cm}
    \end{figure}

\textit{Branch 2. Attack mechanism simulation}. We try to simulate an external attacker in preparation for safety testing in the second branch. Specifically, we follow previous MIA work~\cite{ResAdv} and simulate a black-box MIA attacker with a fully connected neural network, as depicted in Figure~\ref{attack}~(a). The simulated attacker contains three parts: \textit{probability stream}, \textit{label stream}, and \textit{fusion stream}. The probability stream processes the output probability from the target sparse model $\mathcal{M}_{S}(x)$. The label stream deals with the label $y$ of the sample $x$. Then, the fusion stream fuses the features extracted from the two preceding streams and outputs a probability to indicate whether a sample is used in training or not. To improve the efficiency of attacker simulation, we do not independently train an attacker for each candidate sparse model $\mathcal M_s^i$ in Branch 1. Instead, we first train a simulated attacker $\mathcal A$ based on $\mathcal M_S$; then, for each $\mathcal M_s^i$, we fine-tune $\mathcal A$ for multiple epochs to ensure its attack effectiveness toward $\mathcal M_s^i$. This strategy of training acceleration is sensible as $\mathcal M_s^i$ often deviates from $\mathcal M_s$ marginally.

Finally, four sparse model variants (Branch 1) and  corresponding well-trained simulated attackers (Branch 2) are sent to the next stage for safety testing.

\textbf{Stage 3. Safety Test-driven Model Selection.} $\;$ Safety test-driven model selection aims to choose a candidate sparse model with the best trade-off in task performance and safety protection. Specifically, we employ the simulated attacker to conduct safety testing of black-box MIA on four candidate sparse models and then record the attack accuracy. Subsequently, we can simply select the one whose attack accuracy is the lowest (strongest defense ability). However, this manner ignores the task performance, thereby missing a comprehensive consideration. Hence, we also conduct task-performance evaluation on four sparse models (\eg, image classification accuracy for CV tasks). Then, by considering the attack accuracy (safety) and task performance together, we can select the best candidate model. In our implementation, we use a newly defined \textbf{TM-score} (\ie, \textbf{T}ask performance divided by \textbf{M}IA attack accuracy, with details illustrated later) for model selection. The candidate sparse model with the highest TM-score is selected. 
The selected model is then sent back to Stage 2 and a new iteration starts. The whole process will terminate after repeating a predefined number of iterations.

\textbf{Metric.}\label{sub:attack_setup}$\;$ For performance, we evaluate model accuracy on classification tasks, represented by \textit{Task Acc}. The higher the Task Acc is, the better the model performs. For safety, we adopt the accuracy of membership inference attacks, denoted as \textit{MIA Acc}, to reflect defense ability. The lower the MIA Acc is, the stronger the model's defense ability is. Besides, we take both performance and safety into consideration, and design a metric called \textit{TM-score} that aims to directly evaluate the performance-safety trade-off:
\begin{equation}\label{score}
TM\textbf{-}score = \frac{{\quad(Task\; Acc)}^{\lambda}}{\;MIA\; Acc} \,,
\end{equation}
where $\lambda$ is a coefficient to control this trade-off and we set $\lambda=1$ in our approach for simplicity if not specified. 
If a model keeps high Task Acc and low MIA Acc, the TM-score will be high. 
In brief, this metric is in line with our goal to seek a model with both good performance and strong safety.

\textbf{Enhancement with Adversarial Training.} A target model and adversarial models (both usually involved in adversarial training) compete mutually during training. In this way, the target model can improve its capability against adversarial models. Following this idea, we regard the external attacker as the adversarial model and leverage adversarial training based on BMIA-SafeCompress to further improve the target model’s defense ability. Specifically, we additionally construct an adversarial model, \ie, a black-box MIA model $f^{B}_{A}$ following the structure of Figure~\ref{attack} (a). The attack gains are denoted as $G_{f^{B}_{A}}$. Thus, we formalize joint membership privacy attack (gain) and performance objectives in the following min-max optimization problem:
\begin{equation}
\min_{\mathcal{F}} \left \{ \mathcal{L} (\mathcal{F}) + \beta * \max_{f^{B}_{A}}  G_{f^{B}_{A}}(\mathcal{F}) \right \} \\
\end{equation}

$\mathcal{L} (\mathcal{F})$ is our performance objective of BMIA-SafeCompress, where $\mathcal{F}$ is our target sparse model.  $\beta$ is a weighted coefficient. We set it to $0.1$ by default.

\textit{Finally, we may need to point out the difference between the adversarial
model and the simulated attacker in Stage 2 of BMIA-SafeCompress to remove possibly caused confusion}. The simulated attacker in Stage 2 of BMIA-SafeCompress is used for safety testing. They are trained offline and thereby cannot impact the gradient backpropagation when training the target model. In contrast, the adversarial model is trained together with the target model. This process influences the gradient backpropagation and regularizes the target model.

\subsubsection{WMIA-SafeCompress: Defending White-box Membership Inference Attack based on SafeCompress}
Different from black-box MIA, white-box MIA assumes that an adversary has extra knowledge about a target model beyond outputs (\eg, hidden layer parameters~\cite{sablayrolles2019white} and gradient descents in training epochs~\cite{nasr2019comprehensive}). Hence, white-box MIA  could be regarded as a stronger attack manner.  In this part, we aim to configure SafeCompress to white-box MIA. We name this instance of SafeCompress framework as \textbf{\textit{WMIA-SafeCompress}}.

\textbf{Configuration.} To implement WMIA-SafeCompress, concretely, we  include the extra information as inputs to simulate a white-box attacker mechanism $\mathcal A$ proposed in previous work~\cite{nasr2019comprehensive, liu2021ml} while maintaining other processes of BMIA-SafeCompress. In other words, the main adjustment is to replace the simulated black-box attacker in BMIA-SafeCompress with a white-box one to participate in the safety testing. This alteration offers a notable advantage in terms of ease and convenience. As illustrated in Figure~\ref{attack}~(b), the simulated attacker contains five parts: \textit{probability stream, loss stream, gradient stream, label stream, and fusion stream.} The first four streams are designed to process four different inputs,  respectively. They are a target sample's predicted probability, loss value (\eg, classification loss), gradients of the parameters of the target model's previous layer, and one-hot encoding of the target sample's true label. The fifth part, as the name suggests, fuses the extracted features from the first four streams and outputs a probability to indicate whether the sample is used in training or not.

\textbf{Metric.} We simply follow the same metric, \ie, TM-score, in BMIA-SafeCompress. 

We also discuss how to configure SafeCompress for other attacks including attribute inference attack and model inversion attack (attack configurability) and task adaptability using semantic segmentation~\cite{zhu2022crf} as illustrated in \textbf{Appendix A}.

\subsection{Defense against Multiple Heterogeneous Attacks}
In a more practical sense, besides a single attack, AI software could face multiple heterogeneous attacks, which can exacerbate privacy leakage. For example, a potential adversary may employ black-box and/or white-box MIAs. In such scenarios, it becomes difficult for a defender to predict the attack that the adversary may adopt, being black-box, white-box, or both. This situation is more intractable than a single attack stated earlier. To alleviate this issue, we are supposed to take into account all possible attacks, namely, both black-box and white-box MIAs in this context, and introduce our solution below. 

\subsubsection{MMIA-SafeCompress: Extend SafeCompress against Multiple Heterogeneous Membership Inference Attacks}
To defend against multiple heterogeneous attacks, we extend our SafeCompress framework. We list two reasons to support its feasibility. 
(1) The first branch of Stage 2 in SafeCompress, \ie, \textit{dynamic sparse update}, is independent as it does not rely on the second branch, \ie, \textit{attack mechanism simulation}\footnote{The second branch may need candidate models derived from the first branch to help implementation, \eg, our BMIA-SafeCompress and  WMIA-SafeCompress. Safety interaction is performed in Stage 3.}. \textbf{This characteristic enables us to add additional attack mechanisms to simulate while causing little influence on the \textit{dynamic sparse update} branch}.  (2) Our SafeCompress framework strictly follows the \textit{attack configurability} principle. \textbf{In other words, the branch of \textit{attack mechanism simulation}  can be configured with various attacks.} Hence, our SafeCompress framework exhibits high flexibility to seamlessly accommodate multiple heterogeneous attacks through the incorporation of additional simulated attack mechanisms in Stage 2, as well as corresponding safety testing in Stage 3. Further, based on the extension, we can implement various instances against different combinations of attacks toward AI software. Below, we implement a concrete instance called \textbf{\textit{MMIA-SafeCompress}} to defend against black-box and white-box MIAs simultaneously.
 
\textbf{Configuration.} To implement MMIA-SafeCompress, the key is to add simulated black-box and white-box attack mechanisms. In fact, we can build MMIA-SafeCompress based on BMIA-SafeCompress or WMIA-SafeCompress as they already have one simulated attack mechanism.  Here, to ease  illustration, we choose to
build MMIA-SafeCompress based on BMIA-SafeCompress. An extra simulated white-box attacker illustrated in Figure~\ref{attack}~(b) is first added in Stage 2, in parallel with the simulated black-box attacker. Then, corresponding white-box MIA testing is added in Stage 3. Consequently, we shall perform two kinds of safety testing, \ie, black-box and white-box MIA testing, to select the best candidate model that will be sent back to Stage 2, starting a new iteration.

\textbf{Metric.} We leverage a straightforward manner by using the weighted sum to combine two metrics\footnote{More advanced strategies, such as bootstrapping according to different attacks' severity and commonality~\cite{han2019f}, may deserve future research.}:
\begin{equation}\label{mscore}
TM\textbf{-}score_{M} = \alpha * TM\textbf{-}score_{B} + (1-\alpha) * TM\textbf{-}score_{W}
\end{equation}

$TM\textbf{-}score_{B}$ is the score of black-box safety testing while $TM\textbf{-}score_{W}$ is the score of white-box safety testing. $\alpha$ is a coefficient to balance the two metrics. We set $\alpha$ to $0.5$ as we deem that these two metrics are equally important.

\section{Experimental Setup}
In this section, we present an overview of five used datasets, five used models, and eight baselines compared in our work.
\subsection{Datasets and Models}
We conduct experiments on five datasets that are representative with high citation count according to Google Scholar and widely used for membership inference in previous work~\cite{Shaow_Learning, melis2019exploiting, liu2021encodermi}. The partition setting of training/test data  follows the previous work that proposed the respective datasets. For each dataset, we select  as the original big model one model that is famous and highly cited in the domain of computer vision or natural language processing.

\textbf{CIFAR10} and \textbf{CIFAR100}~\cite{cifar_10_krizhevsky2009learning} are two benchmark datasets for image classification. Both of them have $50,000$ training images and $10,000$ test images. CIFAR10 has $10$ categories while CIFAR100 has $100$ categories. The size of every image is $32 \times 32$ pixels. We adopt the AlexNet model~\cite{alexnet_krizhevsky2012imagenet} for CIFAR10 and the VGG16 model~\cite{vgg_simonyan2014very} for CIFAR100.

\textbf{Tiny ImageNet}~\cite{tinyimagenet_le2015tiny} is an image dataset that contains $200$ categories. Each category includes $500$ training images and $50$ test images. The size of each image is $64 \times 64$ pixels. We adopt the  ResNet18 model~\cite{resnet_he2016deep} for Tiny ImageNet.

\textbf{AG News}~\cite{del2005ranking} is a topic classification dataset that has 4 categories. For each category, AG News contains 30,000 training and 1,900 test texts. We adopt the RoBERTa model~\cite{roberta_liu2019roberta} for AG News.

\textbf{Yelp-5}~\cite{Yelpzhang2015character} is a review dataset for  sentiment classification (5 categories). Yelp-5 includes $130,000$ training and $10,000$ test texts per category. We adopt the BERT model~\cite{BERT_devlin2018bert} for Yelp-5.

AlexNet~\cite{alexnet_krizhevsky2012imagenet}, VGG16~\cite{vgg_simonyan2014very}, and ResNet18~\cite{resnet_he2016deep} are convolution neural networks (CNN). BERT~\cite{BERT_devlin2018bert} and RoBERTa~\cite{roberta_liu2019roberta} are transformer-based neural networks. 

\textbf{Dataset Splits for Attacker Simulation.} We split experimental datasets as shown in Table~\ref{data_split} to evaluate a model's defense ability. Following previous work~\cite{Pruning_IJCAI, kd_membership}, we assume that a simulated attacker knows 50\% of a (target) model's training data and 50\% of test data (non-training data). These known data are used for training the attack model. The other data are adopted for evaluation. We put more details of training in \textbf{Appendix B}.

\begin{table}
	\caption{Number of samples in dataset splits.}
	\vspace{-0.3cm}
	\centering	
	\label{data_split}
	\begin{tabular}{l c  c | c  c}
		\toprule
		\multirow{2}{*}{\textbf{Datasets}}&
		\multicolumn{2}{c}{\textbf{Attack Training}}&\multicolumn{2}{c}{\textbf{Attack Evaluation}}\cr
		\cmidrule(lr){2-3} \cmidrule(lr){4-5}
		& $D^{known}_{train}$& $D^{known}_{test}$ &
		$D^{unknown}_{train}$ & $D^{unknown}_{test}$\\
		\midrule
		\textit{CIFAR10} & 25,000 & 5,000 & 25,000 & 5,000 \\
		\textit{CIFAR100} & 25,000 & 5,000 & 25,000 & 5,000 \\
		\textit{Tiny ImageNet} & 50,000 & 5,000 & 50,000 & 5,000 \\ 
		\midrule
		\textit{AG News} & 60,000 & 3,800 & 60,000 & 3,800 \\
		\textit{Yelp-5} & 325,000 & 25,000 & 325,000 & 25,000 \\ 
		\bottomrule
	\end{tabular}
\vspace{-0.2cm}
\end{table}
\setlength{\tabcolsep}{0.2cm}\begin{table}[t]
	\caption{Two-step baseline approaches.}
	\centering
	\vspace{-0.3cm}
	\label{baseline_}
	\begin{tabular}{l c  c | c  c c}
		\toprule
		\multirow{2}{*}{\textbf{Baselines}}&
		\multicolumn{2}{c}{\textbf{Model Compression}}&\multicolumn{3}{c}{\textbf{MIA Defense}}\cr
		\cmidrule(lr){2-3} \cmidrule(lr){4-6}
		& \textit{Pruning} & \textit{KD} &
		\textit{DP} & \textit{AdvReg} & \textit{DMP} \\
		\midrule
		\textit{Pr-DP} & \Checkmark & & \Checkmark &  & \\
		\textit{Pr-AdvReg} & \Checkmark &  & & \Checkmark &\\
		\textit{Pr-DMP} & \Checkmark &  &  &  & \Checkmark\\ 
		\textit{KD-DP} & &\Checkmark & \Checkmark&  & \\ 
		\textit{KD-AdvReg} &  & \Checkmark & & \Checkmark  & \\ 
		\textit{KD-DMP} & & \Checkmark  &  &  &\Checkmark \\ 
		\bottomrule
	\end{tabular}
\vspace{-0.3cm}
\end{table}
\subsection{Baselines} \label{baseline}
Given that no existing studies focus on safe model compression, we  formulate multiple \textit{two-step} baseline approaches by combining state-of-the-art model compression and MIA defense techniques.

In the first step, we  choose one \textbf{compression} technique  $\mathcal{C}$. We consider two widely used and effective choices to compress big models, \ie, \textit{pruning}~\cite{pruning_han2015deep} and \textit{knowledge distillation (KD)}~\cite{KD_hinton2015distilling}. For pruning, we adopt the `pretrain$\to$prune$\to$fine-tune' paradigm. We first pretrain a big model from scratch. Afterward, we leverage magnitude-based pruning and compress this full model to a certain sparsity. Finally, we fine-tune the pruned model to recover model performance. 
For KD, we design a small dense model as a student model given the sparsity requirement. Then, we use a well trained big model as a teacher model to help the student model's training.

In the second step, we select one MIA \textbf{defense} technique $\mathcal{D}$. We test three defense techniques including \textit{differential privacy (DP)}~\cite{dp_abadi2016}, \textit{adversary regularization (AdvReg)}~\cite{ResAdv}, and \textit{distillation for membership privacy (DMP)}~\cite{kd_membership}. DP adds noise to gradients while training a target model. AdvRes adds membership inference gain $G_A$ of an attacker to the loss function of a target model as a regularization term, forming a min-max game.  
DMP first trains an unprotected (vanilla) teacher model. Then, DMP utilizes extra unlabeled data and adopts knowledge distillation to force a student model \ie, a target model, to simulate the output of the teacher. 

Finally, we group two-step techniques by enhancing the training or fine-tuning process in $\mathcal{C}$ with the defense technique in $\mathcal{D}$. For example, the Pr-DP baseline adopts the `pretrain$\to$prune$\to$DP-based fine-tune' process; KD-AdvReg trains a student (a compressed model) with the help of a big teacher model in a min-max game manner. All the six two-step baseline approaches are listed in Table~\ref{baseline_}.

\textbf{Baseline 7 (MIA-Pr).}$\:$ MIA-Pr is originally proposed 
 by previous work~\cite{Pruning_IJCAI} to defend MIA, and its main technique is pruning. Hence, although the previous work~\cite{Pruning_IJCAI} does not specify the usage for model compression, MIA-Pr can naturally reduce a model's size. We use MIA-Pr as a baseline to prune a big model until satisfying a given sparsity requirement. Note that when pruning the big model, MIA-Pr optimizes the defense effect of MIA, while the two-step baseline Pr-X (the first step is pruning) optimizes the task performance (\eg, classification accuracy).

\textbf{Baseline 8 (KL-Div~\cite{pruning_defeat_yuan2022membership}).} KL-Div is initially proposed by previous work~\cite{pruning_defeat_yuan2022membership} to defend against a special attacker that leverages potential defects in pruning. The main idea is to use KL-divergence to align the posterior predictions of different input samples, aiming at improving a compressed model's defense ability. However, this approach does not consider optimizing model performance. In this work, we also regard this approach  as our baseline and follow the setting stated in the previous work~\cite{pruning_defeat_yuan2022membership}.

\section{Experimental Results}
\label{sec:experiemnts}
In this section, we present the results of our BMIA-SafeCompress, WMIA-SafeCompress, and MMIA-SafeCompress, respectively, to show the effectiveness of our SafeCompress framework.
\subsection{Experimental Results on BMIA-SafeCompress}
To show the effectiveness of BMIA-SafeCompress, we conduct experiments on CV and NLP datasets, respectively.
\subsubsection{Results on CV Datasets} \label{comparison_baselines_cv}
\begin{figure}[t]
	\centering{\includegraphics[width=1\linewidth]{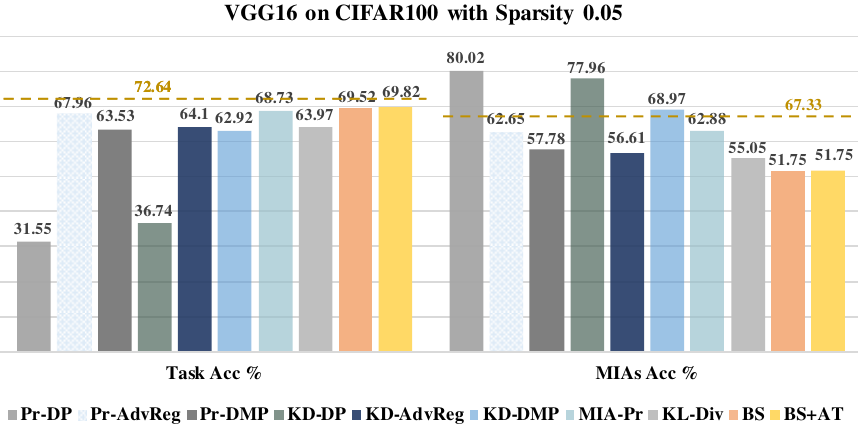}}
	\vspace{-0.7cm}
	\caption{Results on CIFAR100 with sparsity 0.05 (BS: BMIA-SafeCompress, AT: Adversarial Training, dash line: uncompressed VGG16).}
	\label{vgg_0.05}
	\vspace{-0.25cm}
\end{figure}

\begin{figure*}
	\centering{\includegraphics[width=1\linewidth]{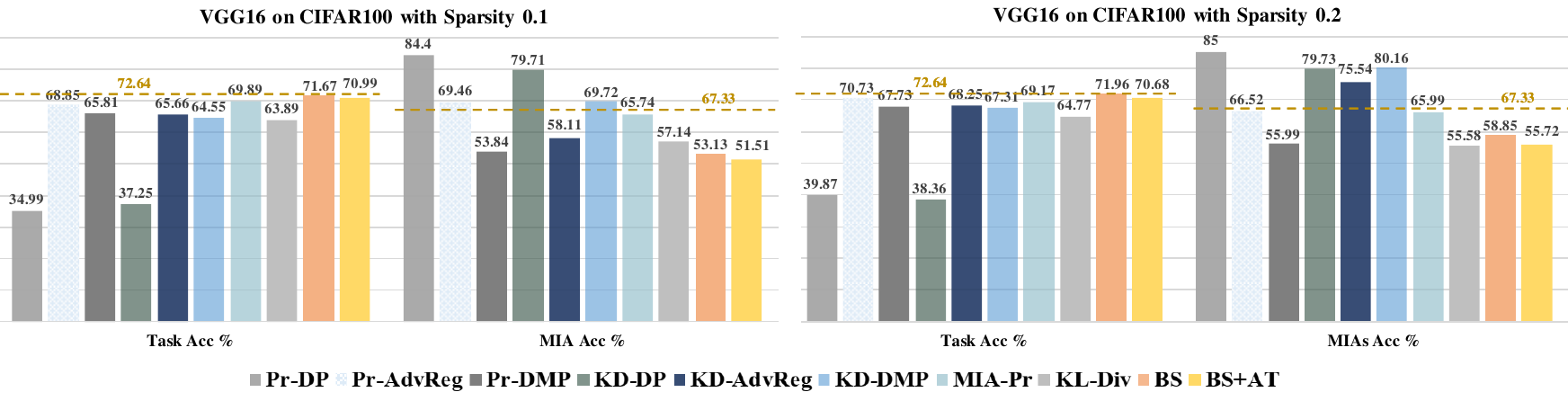}}
	\vspace{-0.8cm}
	\caption{Results on CIFAR100 with sparsity 0.1 and 0.2 (BS: BMIA-SafeCompress, AT: Adversarial Training, dash line: uncompressed VGG16).}
	\label{vgg_0.1_0.2}
	\vspace{-0.3cm}
\end{figure*}

\textbf{CIFAR100 (VGG16).}  We first report Task Acc and MIA Acc on CIFAR100 with sparsity 0.05 using VGG16. As illustrated in Figure~\ref{vgg_0.05}, the results indicate that BMIA-SafeCompress outperforms eight baselines in Task Acc while alleviating MIA risks remarkably. Specifically, BMIA-SafeCompress produces 51.75\% for MIA Acc (just a bit higher than random guess), decreasing by 28.27\%, 10.9\%, 6.03\%, 26.21\%, 4.86\%, 17.22\%, 11.13\%, and 3.30\% compared with eight baselines, respectively.  Moreover, equipping BMIA-SafeCompress with Adversarial Training leads to an additional improvement of 0.30\% in Task Acc while not increasing privacy risk. 
Interestingly, when compared to the uncompressed VGG16, SafeCompress reduces MIA risks to a large extent while sacrificing Task Acc only a little. This result highlights the practicality of SafeCompress in generating a small and safe model with competitive task performance.

Further, to validate the effectiveness on different sparsity requirements, we report the results in Figure~\ref{vgg_0.1_0.2} with sparsity 0.1 and 0.2. Consistent with the results of sparsity 0.05, BMIA-SafeCompress still beats all the baselines by achieving higher Task Acc and lower MIA Acc.
We also observe that when sparsity increases (\ie, the number of non-zero parameters increases), Task Acc of BMIA-SafeCompress increases from 69.52\% (sparsity 0.05) to 71.96\% (sparsity 0.2); meanwhile, MIA Acc also goes up from 51.75\% (sparsity 0.05) to 58.85\% (sparsity 0.2), indicating that the compressed model becomes more vulnerable with increasing sparsity. Such a phenomenon may be incurred by the extra information (\ie, more parameters) kept in the compressed model when sparsity rises --- some of the 
extra information may be generalizable so that task performance is enhanced; other information might be specific to training data, thus leading to higher MIA risks. 

Finally, to gain insight into the performance-safety trade-off in one shot, we report TM-score (Task Acc divided by MIA Acc, Eq.~\eqref{score}) in Table~\ref{vgg_tm-score}. BMIA-SafeCompress (+ Adversarial Training) outperforms all the baselines significantly and consistently under three sparsity settings by making a better trade-off between performance and safety.

\begin{figure}[t]
        \vspace{-0.3cm}
\centering{\includegraphics[width=1\linewidth]{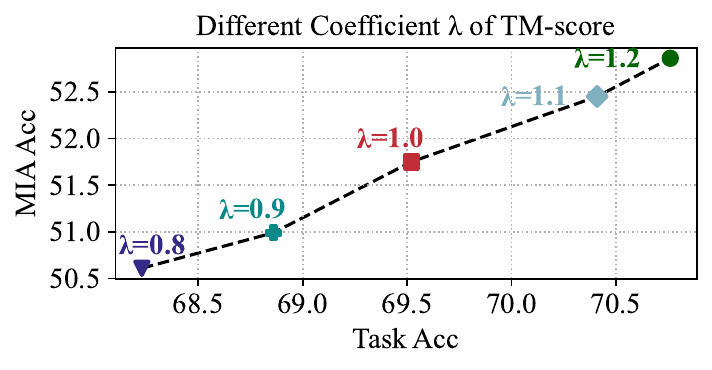}}
	\vspace{-1.0cm}
	\caption{Different coefficient $\lambda$ on BMIA-SafeCompress with sparsity=0.05.}
	\label{fig:Coefficient}
	\vspace{-0.1cm}
\end{figure}

In practice, users may have different requirements for safety and performance. Fortunately, it can be easily achieved in our approach by adjusting the coefficient $\lambda$ of TM-score. For example, a user may think that safety is more important than performance, and therefore she can decrease $\lambda$ to a smaller value (\eg, 0.9 or 0.8); otherwise, she can increase $\lambda$ to a bigger value (\eg, 1.1 or 1.2). We illustrate how the coefficient $\lambda$ controls the trade-off in Figure~\ref{fig:Coefficient}. It is seen that a larger $\lambda$ (performance is important) results in higher performance  while a smaller $\lambda$ results in stronger defense capability.

We also explore six other aspects of our framework based on BMIA-SafeCompress in \textbf{Appendix C.1.1}. The first is time consumption where the time consumption of BMIA-SafeCompress is comparable to that of other baselines. We further find that, in BMIA-SafeCompress, attack mechanism simulation consumes more time than that of sparse model training. The second is the rationality of selecting the candidate sparse model that performs the best in the safety testing.  We empirically demonstrate that compared to enumerating all candidate sparse models, this
strategy in SafeCompress is reasonable in light of the performance and significantly reduces time consumption. The third is the flexibility of SafeCompress to incorporate other training tricks where we show that SafeCompress is compatible with most training tricks including dropout, data augmentation, adversary regularization, \etc. The fourth is to perform SafeCompress on a bigger model (\eg, ResNet50~\cite{resnet_he2016deep}) and a larger dataset (\eg, ImageNet~\cite{deng2009imagenet}). The result indicates that our framework is scalable. The fifth is the comparison between BMIA-SafeCompress and baselines on TM-scores with different $\lambda$ where SafeCompress is shown  to be robust against metric alteration. The last is a statistical analysis with multiple runs, showing that our approach is statistically superior to others. Besides, we conduct experiments on CIFAR10 and Tiny ImageNet in \textbf{Appendix C.1.2} where the superior results over other baselines further show the effectiveness of BMIA-SafeCompress.

\setlength{\tabcolsep}{0.45cm}\begin{table}[t]
	\caption{Defend against black-box MIA. TM-score$_{B}$ on CIFAR100.  The best results are marked in \underline{\textbf{bold}}.}
	\vspace{-0.3cm}
	\label{vgg_tm-score}
	\begin{tabular}{l  c  c  c}
		\toprule
		\multirow{2}{*}{\textbf{Approach}}&
		\multicolumn{3}{c}{\textbf{Sparsity}} \cr
		\cmidrule(lr){2-4}
		& 0.05 & 0.1 & 0.2 \\
		\midrule
		\textit{VGG16} (uncompressed) & 1.08 & 1.08 & 1.08  \\
		\midrule
		\textit{Pr-DP} & 0.39 & 0.41 & 0.47  \\
		\textit{Pr-AdvReg} & 1.08
		& 0.99 & 1.06 \\
		\textit{Pr-DMP} & 1.10
		& 1.22 & 1.21 \\
		\textit{KD-DP} & 0.47 & 0.47 & 0.48 \\
		\textit{KD-AdvReg} & 1.13 & 1.13 & 0.90 \\
		\textit{KD-DMP} & 0.91 & 0.93 & 0.84 \\
		\textit{MIA-Pr} & 1.09 & 1.06 & 1.05 \\
		\textit{KL-Div} & 1.16 & 1.12 & 1.17 \\
		\midrule
		\textbf{\textit{BMIA-SafeCompress}} & 1.34 & 1.35 & 1.22 \\
		\textbf{\textit{+ Adversarial Training}} & $\underline{\textbf{1.35}}$  &  $\underline{\textbf{1.38}}$ & $\underline{\textbf{1.27}}$ \\
		\bottomrule
	\end{tabular}
\vspace{-0.3cm}
\end{table}

\setlength{\tabcolsep}{0.2cm}\begin{table}[t]
	\centering
	\vspace{-0.1cm}
	\caption{Defend against black-box MIA. Task Acc (performance), MIA Acc (safety), and TM-score$_{B}$ on AG News.}
	\vspace{-0.3cm}
	\label{roberta_0.5}
	\begin{tabular}{l  c  c  c}
		\toprule
		\multirow{2}{*}{\textbf{Approach}}&
		\multicolumn{3}{c}{\textbf{Sparsity=0.5} }\cr
		\cmidrule(lr){2-4}
		& \textit{Task Acc} & \textit{MIA Acc} & \textit{TM-score}
		\\
		\midrule
		\textit{RoBERTa} (Uncompressed) & 89.20\% & 57.59\% & 1.55 \\
		\midrule
		\textit{Pr-DP} & 87.28\% & 56.07\% & 1.56 \\
		\textit{Pr-AdvReg} & 87.24\% & 57.58\% & 1.52 \\
		\textit{Pr-DMP} & 86.38\% & 56.31\% & 1.53 \\
		\textit{KD-DP} & 82.64\% & 56.13\% & 1.47 \\
		\textit{KD-AdvReg} & $\underline{\textbf{88.10\%}}$ & 56.56\% & 1.56 \\
		\textit{KD-DMP} & 87.24\% & 56.64\% & 1.54 \\
		\textit{MIA-Pr} & 87.91\% & 57.04\% & 1.54 \\
		\textit{KL-Div} &  83.26\% & $\underline{\textbf{54.09\%}}$ &  1.54\\
		\midrule
		\textit{\textbf{BMIA-SafeCompress}} & 87.50\% & 55.28\% &  1.58 \\
		\textit{\textbf{+ Adversarial Training}} & 87.63\% &  54.77\% &$\underline{\textbf{1.60}}$ \\
		\bottomrule
	\end{tabular}
\vspace{-0.3cm}
\end{table}
\setlength{\tabcolsep}{0.45cm}\begin{table*}[h]
	\centering
	\caption{Defend against white-box MIA. Task Acc (performance), MIA Acc (safety) and TM-score$_{W}$ results on CIFAR100. The best results are marked in \underline{\textbf{bold}}.}
	\vspace{-0.3cm}
	\label{vgg16_0.05_0.1_w}
	\begin{tabular}{l  c  c  c c c  c  c}
		\toprule
		\multirow{2}{*}{\textbf{Approach}}&
		\multicolumn{3}{c}{\textbf{Sparsity=0.05} }& &\multicolumn{3}{c}{\textbf{Sparsity=0.1}}\cr
		\cmidrule(lr){2-4} \cmidrule(lr){6-8}
		& \textit{Task Acc} & \textit{MIA Acc} & \textit{TM-score} &
		& \textit{Task Acc} & \textit{MIA Acc} & \textit{TM-score} \\
		\midrule
		\textit{VGG16} (uncompressed) & 72.64\% & 71.58\% & 1.01  & & 72.64\% & 71.58\% & 1.01 \\
		\midrule
		\textit{Pr-DP} & 31.55\% & 84.18\% & 0.37 & & 34.99\% & 82.73\% & 0.42 \\
		\textit{Pr-AdvReg} & 67.96\% & 67.05\% & 1.01 & & 68.85\%  & 70.65\% &  0.97 \\
		\textit{Pr-DMP} & 63.53\% & $\underline{\textbf{54.01\%}}$ & 1.18 & & 65.81\% & $\underline{\textbf{52.28\%}}$ & 1.26 \\
		\textit{KD-DP} & 36.74\% & 77.40\% & 0.47 & & 37.25\% & 78.17\% & 0.48  \\
		\textit{KD-AdvReg} &  64.10\% & 61.27\% & 1.05 & & 65.66\% & 63.87\% & 1.03  \\
		\textit{KD-DMP} &  62.92\% & 63.30\% & 0.99 & & 64.55\% & 61.67\% & 1.05 \\
		\textit{MIA-Pr} & $\underline{\textbf{68.73\%}}$ & 66.77\% & 1.03 & & 69.89\%& 68.59\% &  1.02 \\
		\textit{KL-Div} & 63.97\% & 56.54\% & 1.13 & & 63.89\% & 57.73\% & 1.11 \\
		\midrule
		\textbf{\textit{WMIA-SafeCompress}} & 67.51\% & 56.22\% & 1.20  & & $\underline{\textbf{70.53\%}}$ & 53.44\%  &$\underline{\textbf{1.32}}$  \\
		\textbf{\textit{+ Adversarial Training}} & $\underline{\textbf{68.73\%}}$ & 55.63\% & $\underline{\textbf{1.24}}$ & &70.51\% & 53.25\% & $\underline{\textbf{1.32}}$ \\
		\bottomrule
	\end{tabular}
\vspace{-0.3cm}
\end{table*}

\subsubsection{Results on NLP Datasets}
\textbf{AG News (RoBERTa).} To validate the effectiveness and generalization of BMIA-SafeCompress, we conduct experiments on AG News. As indicated in Table~\ref{roberta_0.5}, BMIA-SafeCompress achieves 87.50\% for Task Acc, slightly inferior to the highest 88.10\% (KD-AdvReg). However, our approach decreases MIA Acc to 55.28\%, much lower than KD-AdvReg (56.56\%). BMIA-SafeCompress outperforms all the baselines in MIA Acc, resulting in a high TM-score (1.58). Moreover, incorporating BMIA-SafeCompress with Adversarial Training leads to a further enhancement in all the metrics. The results on Yelp dataset in \textbf{Appendix C.1.3} further show the superiority of BMIA-SafeCompress over other baselines and demonstrate its effectiveness. 

\subsection{Experimental Results on WMIA-SafeCompress}
\textbf{CIFAR100 (VGG16).} The results are reported in Table~\ref{vgg16_0.05_0.1_w}. When model sparsity is set to 0.05, our approach, namely WMIA-Safecompress, produces 67.51\% for Task Acc, achieving a competitive classification accuracy among all the baselines. In addition, our approach also decreases MIA Acc by 15.36\% compared to the uncompressed VGG16. Although it is a little bit inferior (2.21\% lower) in MIA Acc than Pr-DMP,  our approach maintains higher performance (3.98\% higher) than Pr-DMP. We also calculate the TM-score for each approach to show its trade-off degree. It is observed that our approach obtains the highest TM-score (1.20). Similar results can be observed for sparsity 0.1 where WMIA-Safecompress also achieves the highest TM-score (1.32). In total, this result is reasonable as our framework, SafeCompress, targets a bi-objective (safety and performance) optimization process.
Simultaneously, this result indicates that our framework can be flexibly configured for other, even stronger attacks. Leveraging  Adversarial Training leads to a further descent in privacy risk, especially for sparsity 0.05. 

Besides CIFAR100, the experimental results  on CIFAR10 and Tiny ImageNet (CV) in \textbf{Appendix C.2.1} and AG News  (NLP) in \textbf{Appendix C.2.2} show the effectiveness of WMIA-SafeCompress.

\subsection{Experimental Results on MMIA-SafeCompress}
\textbf{CIFAR100 (VGG16).} As shown in Table~\ref{vgg16_0.05_0.1_m}, our approach, namely MMIA-SafeCompress, achieves the most competitive classification performance in both sparsity settings. Specifically, it produces 68.13\% (sparsity=0.05) for Task Acc, just 0.60\% lower than MIA-Pr (68.73\%). When sparsity is 0.1, it obtains 70.48\%, outperforming all the baselines in Task Acc. At the same time, it holds the lowest MIA Acc$_B$ among all the approaches under comparison. For MIA Acc$_W$, although our approach is a little bit inferior to Pr-DMP, our approach obtains relatively low MIA Acc$_W$ in both sparsity settings compared to the defense effects of the baselines. Finally, our approach  produces 1.23 (sparsity=0.05) and 1.25 (sparsity=0.1) for TM-score, outperforming all the baselines. The results indicate that our approach is able to handle heterogeneous membership inference attacks and make the performance-safety trade-off, demonstrating the feasibility of extending our SafeCompress framework against heterogeneous attacks.

When our approach is equipped with Adversarial Training, MIA Acc$_B$ further decreases in both sparsity settings. We also observe that Adversarial Training has an inconsistent influence on Task Acc and MIA Acc$_W$ for different sparsity settings. Specifically, Adversarial Training improves Task Acc when sparsity is 0.05 and brings a slight descent when sparsity is 0.1. With Adversarial Training, MIA Acc$_W$ increases when sparsity is 0.05 and decreases when sparsity is 0.1. This phenomenon is interesting. Since it is not our main interest, we leave it for future exploration. Adversarial Training further brings improvements to TM-Score for both sparsity settings, showing its effectiveness.

Another observation is that our approach  seems to pay more attention to defending against black-box MIA than white-box MIA. We speculate that white-box MIA is harder to defend against than black-box one as white-box MIA involves more information (\eg, hidden layer parameters~\cite{sablayrolles2019white} and gradient descents in training epochs~\cite{nasr2019comprehensive}). So when the two heterogeneous attackers attack our model simultaneously, our approach is inclined to defend against the black-box attacker more. It could be an advantage to some extent as the black-box attack is easier to perform compared to the white-box one, which needs to collect extra information. Similar situations can be seen in some  baselines such as KD-AdvReg and MIA-Pr. We consider that this phenomenon may be general and it is interesting to find the potential reason behind it in the future.

Besides CIFAR100, in \textbf{Appendix C.3.1} and \textbf{Appendix C.3.2}, we also show the performance of our approach on CIFAR10, Tiny ImageNet, and AG News, respectively. These results further indicate the effectiveness and generalization of SafeCompress in defending against multiple heterogeneous attacks.

\setlength{\tabcolsep}{0.2cm}\begin{table*}[h]
	\centering
	\caption{Defend against black-box and white-box MIAs. Task Acc (performance), MIA Acc (safety) and TM-score$_{M}$ results on CIFAR100. The best results are marked in \underline{\textbf{bold}}.}
	\vspace{-0.3cm}
	\label{vgg16_0.05_0.1_m}
	\begin{tabular}{l  c  c  c c  c  c c  c  c}
		\toprule
		\multirow{2}{*}{\textbf{Approach}}&
		\multicolumn{4}{c}{\textbf{Sparsity=0.05} }& &\multicolumn{4}{c}{\textbf{Sparsity=0.1}}\cr
		\cmidrule(lr){2-5} \cmidrule(lr){7-10}
		& \textit{Task Acc} & \textit{MIA Acc$_{B}$} & \textit{MIA Acc$_{W}$} & \textit{TM-score} &
		& \textit{Task Acc} & \textit{MIA Acc$_{B}$} & \textit{MIA Acc$_{W}$} & \textit{TM-score} \\
		\midrule
		\textit{VGG16} (uncompressed) & 72.64\% & 67.33\% & 71.58\% & 1.05 &  & 72.64\% & 67.33\% & 71.58\% & 1.05 \\
		\midrule
		\textit{Pr-DP} & 31.55\% & 80.02\% & 84.18\% & 0.38 & & 34.99\% & 84.40\% & 82.73\% & 0.42 \\
		\textit{Pr-AdvReg} & 67.96\% & 62.65\% & 67.05\% & 1.05 & & 68.85\%  & 69.46\% & 70.65\% & 0.98  \\
		\textit{Pr-DMP} & 63.53\% & 57.78\% & $\underline{\textbf{54.01\%}}$ & 1.14 & & 65.81\% & 53.84\%& $\underline{\textbf{52.28\%}}$ & 1.24 \\
		\textit{KD-DP} & 36.74\% & 77.96\% & 77.40\% & 0.47 & & 37.25\% & 79.71\% & 78.17\% & 0.47 \\
		\textit{KD-AdvReg} & 64.10\% & 56.61\% & 61.27\% & 1.09 & & 65.66\% & 58.11\% & 63.87\% & 1.08 \\
		\textit{KD-DMP} & 62.92\% & 68.97\% & 63.30\% & 0.95 & & 64.55\% & 69.72\% & 61.67\% & 0.99 \\
		\textit{MIA-Pr} & 68.73\% & 62.88\% & 66.77\% & 1.06 & & 69.89\%& 65.74\% & 68.59\% & 1.04 \\
		\textit{KL-Div} & 63.97\% & 55.05\% & 56.54\% & 1.15 & & 63.89\% & 57.14\% & 57.73\% & 1.11 \\
		\midrule
		\textbf{\textit{MMIA-SafeCompress}} &  68.13\% & 52.32\% & 59.01\% & 1.23 & & $\underline{\textbf{70.48\%}}$  & 51.79\%  & 61.60\% & 1.25 \\
	     \textbf{\textit{+ Adversarial Training}} & $\underline{\textbf{69.17\%}}$ & $\underline{\textbf{51.23\%}}$ & 61.17\% & $\underline{\textbf{1.24}}$ &  & 70.31\% & $\underline{\textbf{51.66\%}}$  & 58.66\% &  $\underline{\textbf{1.28}}$\\
		\bottomrule
	\end{tabular}
\vspace{-0.5cm}
\end{table*}

\section{Related Work}
In this section, we discuss related work from three aspects including membership inference attack,  membership inference defense, and model compression.
\subsection{Membership Inference Attack}
Membership inference attack (MIA)~\cite{Shaow_Learning, sablayrolles2019white}, which aims to infer whether a data record is used to train a model or not, has the potential to raise severe privacy risks to individuals.  A prevalent attack fashion is to train a neural network via multiple shadow training~\cite{Shaow_Learning}, converting the task of recognizing member and non-member of training datasets to a binary classification problem. Unlike binary classifier-based MIA, another attack form, metric-based MIA that just computes metrics (\eg, prediction correctness~\cite{dp_yeom2018privacy} or confidence~\cite{pre_conf_ML_Leaks}, and entropy~\cite{song2021systematic_entropy}) is simpler and less computational. However, the performance of metric-based attacks is inferior to that of classifier-based attacks. Liu~\etal~\cite{liu2022membership} exploit membership information from the whole training process of a target model for improving attack performance. Recently, membership inference attack has been widely extended to more fields. For example, MIA is investigated on generative
models~\cite{hayes2017logan} and mainly focuses on GANs~\cite{goodfellow2020generative}. Song and Raghunathan~\cite{song2020information} introduce the first MIA
on word embedding and sentence embedding models that map
raw objects (\eg, words, sentences, and graphs) to real-valued
vectors. Duddu~\etal~\cite{duddu2020quantifying} introduce the first
MIA on graph embedding models. Tseng~\etal~\cite{tseng2021membership} present
the first MIA on self-supervised speech models under
black-box access. M4I~\cite{hu2022m} is the first to investigate membership inference on multi-modal models. Liu~\etal~\cite{liu2021encodermi} propose the
first membership inference approach called EncoderMI
for contrastive-learning based models.

\subsection{Membership Inference Defense}
Considering that membership inference attack has the potential to raise severe privacy risks to individuals, studying how to defend against it becomes
important. For example, differential privacy~\cite{dp_abadi2016, dp_rahman2018} (known as DP) interrupts an attack model by adding noise to the learning object or output of a target model. But the cost between utilization and defense is unacceptable~\cite{dp_jayaraman2019evaluating_unaccept}. To improve the model utility,  Jia~\etal~\cite{memguard_jia2019} introduce Memguard, which  adds carefully computed noise to the output of a target model, aiming to defend an attack model while keeping performance. However, Memguard is vulnerable to threshold-based attack~\cite{song2021systematic_entropy}. Adversarial regularization~\cite{ResAdv}, known as AdvReg, combines a target model's training process with that of an attack model, formulating a min-max game. Yang~\etal~\cite{yang2022purifier} propose Purifier to defend inference attacks via transforming the vectors of confidence score predicted by a target classifier. More recently, model compression technologies (\eg, knowledge distillation~\cite{KD_hinton2015distilling} and pruning~\cite{pruning_han2015deep}) have been employed to protect member privacy. Based on knowledge distillation, Shejwalkar and Houmansadr~\cite{kd_membership} propose Distillation For Membership Privacy (DMP) defense, where a teacher is trained to label an unlabeled reference dataset, and those with low prediction entropy are selected to train a target model. Further, Zheng~\etal~\cite{MIAs_KD_V2} propose complementary knowledge distillation (CKD) and pseudo-complementary knowledge distillation (PCKD). Tang~\etal~\cite{tang2022mitigating} propose to leverage distillation to distill a training dataset via a novel ensemble architecture. However, as knowledge distillation is an indirect learning strategy~\cite{Moonshine_NEURIPS2018_49b8b4f9}, some critical information may be lost while mimicking the teacher. In more recent work~\cite{Pruning_IJCAI}, pruning is adopted to mitigate MIA while reducing model size simultaneously. But that work mainly focuses on preventing membership inference attack without explicitly considering the memory restriction. 

Our work  aims to address the problem of safe model compression, being  different from the preceding previous work, which  mainly focuses on defending against MIA. More specifically, our work aims to decrease the risk of privacy attacks (\eg, MIA) and  keep excellent performance when compressing a deep neural model (\eg, DNN). Essentially, the target problem in our work is a bi-objective optimization problem. 

\subsection{Model Compression}
Due to limited memory and computation resources,  
model compression~\cite{pruning_han2015deep} plays a crucial role, especially when transformer-based big models~\cite{BERT_devlin2018bert,VIT_dosovitskiy2020vit} become the mainstream. To alleviate the issue, various approaches have been proposed. For example, pruning~\cite{pruning_han2015deep}, as a direct and effective approach, removes unimportant weights or structures according to certain criteria (\eg, weight magnitude). Knowledge distillation~\cite{KD_hinton2015distilling}, known as KD, transforms knowledge from a big model (named as a teacher) to a small model (named as a student) during training. Quantification~\cite{quantify_han2015learning} converts a long storage width in memory to a shorter one, \eg, converting float (64 bit) to 8-bit integer~\cite{8_quantization_jacob2018quantization}, even to binary (1 bit)~\cite{binary_hubara2016binarized}. In addition, dynamic sparse training~\cite{DST_Begin_mocanu2018scalable} (denoted as DST) is proposed as a new compression approach, and  achieves surprising performance, attracting much attention from researchers. Follow-up work further introduces weight redistribution~\cite{weight_redistribut_dettmers2019sparse, anvance_mostafa2019parameter}, gradient-based weight growth~\cite{weight_redistribut_dettmers2019sparse,evci2020rigging}, and extra weights update in the backward pass~\cite{backupdate_jayakumar2020top,backupdate_raihan2020sparse} to improve the sparse training performance. More recently, the DST~\cite{liu2021sparse} paradigm has shown great potential to deploy neural network models into edge devices at a large scale. Lin~\etal~\cite{lin2022device} design an algorithm-system co-design framework to make on-device training possible with only 256KB of memory.
While there exists  numerous work in model compression to balance  size and performance, privacy safety is not well considered. Differing from the preceding work, our work considers both safety and performance during model compression.

\section{Limitations and Future Work}
\label{sec:discussion}

As a study toward safe model compression, we have also identified  a number of  possibilities of future work  that may attract more effort to this direction.

\textbf{Extending SafeCompress to more heterogeneous attacks}. In practice, AI software could face various types of attacks and even their combinations. The SafeCompress framework follows the \textit{attack configurability} principle to be able to be configured against different attacks and has the potential to be extended to defend against heterogeneous attacks. In this work, we consider black-box MIA and white-box MIA, two heterogeneous attacks, to conduct experiments across computer vision and natural language processing. It is interesting and worth expecting to include more heterogeneous attacks in more domains as the SafeCompress framework is flexible and easily extensible.

\textbf{Benchmarking performance-safety trade-offs between models}. In this work, for any dataset, we use only one state-of-the-art model as the original big model. Prior research~\cite{nasr2019comprehensive} has pointed out that, when two models do the same task with similar performance, the model with more parameters may face higher MIA risks. Then, if we compress the two models with the same sparsity restrictions, could the statement still stand or be overturned? Addressing this question could significantly help the model selection process in AI software deployment.

\section{Conclusion}

In this article, we have presented a performance-safety co-optimization framework, called SafeCompress, to address the problem of \textbf{safe model compression}, as it is critical for current large-scale AI software deployment. SafeCompress is a test-driven sparse training framework, which can be easily configured to fight against pre-specified attack mechanisms. Specifically, by simulating the attack mechanisms, SafeCompress performs both safety testing and performance evaluation and iteratively updates the compressed sparse model.
Based on SafeCompress, we have implemented three concrete instances called the BMIA-SafeCompress, WMIA-SafeCompress, and MMIA-SafeCompress approaches against heterogeneous membership inference attacks. Extensive experiments have been conducted on five datasets including three computer vision tasks and two natural language processing tasks. The results demonstrate the effectiveness and generalization of our framework. We have also elaborated on how to adopt SafeCompress to other attacks, incorporate other training tricks into BMIA-SafeCompress, perform SafeCompress on a bigger model and a larger dataset, and conduct  robustness analysis on the metric of TM-score, showing the flexibility, scalability, and robustness of SafeCompress. As a study toward safe model compression, we expect that our work can attract more effort to this promising direction in the new era when AI software is increasingly prevalent. 

\section*{Acknowledgment}
We thank the anonymous reviewers for their constructive comments. This work is supported by the NSFC Grants no. 61972008, 72031001, 72071125, and 62161146003, along with the Tencent Foundation/XPLORER PRIZE.

\newpage

\appendices
\section{}
\setlength{\tabcolsep}{0.05cm}\begin{table*}[h]
	\centering
	\caption{Time consumption on CIFAR100. VGG16 is the uncompressed model. Others are compressed with sparsity 0.05.}
	\label{time}
	\begin{tabular}{c  c  c c  c  c c c c c c}
		\toprule
		\textit{Approach} & 
		\textit{VGG16} (uncompressed)  &
		\textit{Pr-DP} &
		\textit{Pr-AdvReg} &
		\textit{Pr-DMP} &
		\textit{KD-DP} &
		\textit{KD-AdvReg} &
		\textit{KD-DMP} &
		\textit{MIA-Pr} &
		\textit{KL-Div} &
		\textit{\textbf{BMIA-SafeCompress}}
		\\
		\midrule
		\textit{Time} (h) & 1.05   & 16.90  & 2.92
		& 4.75 & 4.89
		
		& 2.22 & 1.42 & 2.37 & 2.27 & 3.98 \\
		\bottomrule
	\end{tabular}
\end{table*}
\subsection{Configuring SafeCompress to Other Attacks}
\label{sub:configure_other_attacks}

To adapt BMIA-SafeCompress (or WMIA-SafeCompress) to other attacks, the main modification is using another attack mechanism instead of black-box (or white-box) MIA in Branch 2 of Stage 2. Two examples are listed as follows.

\textbf{Attribute Inference Attack.} Attribute inference attack (AIA) aims to infer private information of a user (sample), such as age and location~\cite{han2019f}. Suppose that a big model (\eg, auto-encoder~\cite{bengio2013representation}) is trained to obtain a representation from a user's movie ratings to serve downstream tasks such as recommendation; this representation may leak the user's private information~\cite{kosinski2013private}. Then, we can simulate an attacker $\mathcal A$ as a specific AIA approach (\eg, logistic regression~\cite{kosinski2013private}), or a basket of multiple AIA approaches~\cite{han2019f}. Note that the safety and performance metrics also need to be modified. For safety, the metric depends on the private attribute --- \eg, to protect age inference, the metric could be set to mean absolute error. For performance, we can list several downstream tasks and evaluate the learned representations' effects on these tasks.

\textbf{Model Inversion Attack.} ModInv~\cite{fredrikson2015model, zhang2020secret} aims to reconstruct data
samples from a target ML model. In other words, they allow an adversary
to directly learn information about a training dataset. For example, a ModInv adversary could perform such an attack in a facial recognition system, aiming to learn the facial data of a victim whose data is
used to train the model in the system. Then, we can simulate an attacker $\mathcal A$ as a specific ModInv approach (\eg, to synthesize the training dataset relying on GANs~\cite{goodfellow2020generative, zhang2020secret}). Also, the safety and performance metrics need to be modified.

\subsection{Task Adaptability}
SafeCompress can work not only for classification tasks. While classification models are widely used as targets in membership inference attack~\cite{liu2021encodermi, shokri2017membership, salem2018ml, kaya2021does}, we adopt classification task in our work following them. Nevertheless, it is worth emphasizing that SafeCompress can be utilized for other tasks as long as those tasks meet two specific criteria: 1) Big models are built by deep neural networks, \eg, DNN, CNN, RNN, LSTM, Transformer, \etc, to enable dynamic sparse training that is operated on the neural network connection (model weight). 2) The attack mechanism can be simulated by a defender as SafeCompress needs to use the simulated attacker to perform a safety testing. We conduct a preliminary experiment by taking a medical segmentation task for example. We use a representative UNet~\cite{unet} model to perform segmentation on Breast Ultrasound Images Dataset~\cite{al2020dataset}. We set sparsity to 0.1 with a black-box setting and adopt segmentation membership inference~\cite{he2020segmentations} in our experiment. BMIA-SafeCompress obtains $64.67\%$ Task Acc (mIoU) and $52.83\%$ MIA Acc, resulting in $1.22$ of TM-score which is superior to UNet (uncompressed) under the same 100 training epochs whose TM-score is $1.20$ ($68.24\%$ Task Acc and $56.66\%$ MIA Acc).
The result indicates that our framework also can be extended to other tasks.

\section{}
\subsection{Implementation Details}
We perform all the experiments using Pytorch 1.8. on Ubuntu 20.04. For CV tasks, we use NVIDIA  1080Ti with the CPU of Intel Xeon Gold 5118 (4 cores, 2.3GHz) and 16GB memory. We train big models 
with batch size $128$ for $200$ epochs. For NLP tasks, we use NVIDIA 3090 with the CPU of Intel Xeon Gold 5218 (6 cores, 2.3GHz) and  40 GB memory. The batch size is 256 and the training lasts for 10 epochs.  

In SafeCompress, for dynamic sparse training, we simply follow previous work~\cite{DST_InTime} and use its default hyperparameters, \eg, SGD optimizer, learning rate 0.1, and multi-step decay strategy for the learning rate. We only prolong the training epochs from 250 to 300 to help a sparse model enumerate more sparse topology structures. In the fine-tuning process, we set the batch size to $128$ with Adam optimizer whose weight decay is set to $0.05$ and betas is set to $(0.9, 0.999)$ by default. We leverage a small learning rate $5e-4$ and train until it is converged (usually $5 \sim 10$ epochs). To simulate the attacker, following previous work~\cite{ResAdv, sablayrolles2019white, liu2021ml}, we adopt ReLu as the activation function in our attack neural network models. All the network weights are initialized with normal distribution, and all biases are set to 0 by default. The batch size is 128. We use the Adam optimizer with a learning rate of 0.001 and we train the attack models 100 epochs. During the training process, we ensure that every training batch contains the same number of member and non-member data samples, aiming to prevent the attack model from being biased toward either side.

\setlength{\tabcolsep}{0.1cm}\begin{table}[h]
	\centering
	\caption{Using other training tricks in \textit{BMIA-SafeCompress}.}
	\label{flexsibility}
	\begin{tabular}{l  l  l l l l l l}
		\toprule
		\textit{Trick} & 
		\textit{Vanilla} &
		\textit{Dropout} &
		\textit{Aug} &
		\textit{AdvReg} &
		\textit{BigPara} &
		\textit{KD}&
		\\
		\midrule
		\textit{Task Acc} & 69.52\%  & 
		69.43\%
		& 62.03\% & 68.86\% & 68.52\% & $\underline{\textbf{70.02\%}}$ \\ 
		\textit{MIA Acc}& $\underline{\textbf{51.75\%}}$  &
		53.94\% &
		65.46\% & 52.96\% & 52.05\% & 53.10\% \\ 
		\midrule
		\textit{TM-score} & $\underline{\textbf{1.35}}$ &
		1.29 &
		0.95 & 1.30& 1.32 & 1.32 \\ 
		\bottomrule
	\end{tabular}
\end{table}

\setlength{\tabcolsep}{0.22cm}\begin{table}[h]
	\centering
	\caption{Different settings of $\lambda$ for TM-score$_{B}$ on CIFAR100.}
	\label{lambda_tm-score}
	\begin{tabular}{l  c  c  c  c  c}
		\toprule
		\multirow{2}{*}{\textbf{Approach}}&
		\multicolumn{5}{c}{\textbf{Coefficient $\lambda$}} \cr
		\cmidrule(lr){2-6}
		& 0.8 & 0.9 & 1 & 1.1 & 1.2 \\
		\midrule
		\textit{VGG16} (uncompressed) & 0.46 & 0.70 & 1.08 & 1.66 & 2.54 \\
		\midrule
		\textit{Pr-DP} & 0.20 & 0.28 & 0.39 & 0.56  & 0.79  \\
		\textit{Pr-AdvReg} & 0.47 & 0.71 & 1.08
		& 1.65 & 2.52 \\
		\textit{Pr-DMP} & 0.48 & 0.73 & 1.10
		& 1.67 & 2.52 \\
		\textit{KD-DP} & 0.23 & 0.33 & 0.47 & 0.68  & 0.97 \\
		\textit{KD-AdvReg} & 0.49 & 0.75 & 1.13 & 1.72  & 2.60 \\
		\textit{KD-DMP} & 0.40 & 0.60 & 0.91 & 1.38 & 2.09 \\
		\textit{MIA-Pr} & 0.47 & 0.72 & 1.09 & 1.67 & 2.55 \\
		\textit{KL-Div} &  0.51 & 0.77 & 1.16  & 1.76 & 2.67 \\
		\midrule
		\textbf{\textit{BMIA-SafeCompress}} & $\underline{\textbf{0.58}}$&$\underline{\textbf{0.88}}$ & $\underline{\textbf{1.34 }}$ & $\underline{\textbf{2.05}}$ & $\underline{\textbf{3.14}}$ \\
		\bottomrule
	\end{tabular}
\end{table}

\section{}
\subsection{Experimental Results on BMIA-SafeCompress}
\subsubsection{Potential Tapping}

We also explore other six aspects of our framework based on BMIA-SafeCompress. 
\setlength{\tabcolsep}{0.6cm}\begin{table*} [h]
    \caption{Results on CIFAR100 with sparsity 0.05. We repeat three times and report the mean result and standard error.}
    \label{vgg_mutiple_run}
    \centering
    \begin{tabular}{l  c  c  c }
        \toprule
        \textbf{Approach}  & \textbf{Task Acc} & \textbf{MIA Acc} & \textbf{TM-score} \\
        \midrule
        \textit{VGG16} (uncompressed) & $72.70 \pm 0.06$ & $67.87 \pm 0.52$ &  $1.07 \pm 0.0081$ \\
        \midrule
        \textit{Pr-DP} & $31.64 \pm 0.11$ & $80.14 \pm 0.11$ & $0.39 \pm 0.0014$ \\
        \textit{Pr-AdvReg} & $67.93 \pm 0.05$ & $62.75 \pm 0.16$ & $1.08 \pm 0.0023$ \\
        \textit{Pr-DMP} & $63.44 \pm 0.15$ & $57.78 \pm 0.24$ & $1.10 \pm 0.0024$ \\
        \textit{KD-DP} &  $36.81 \pm 0.24$ & $78.08 \pm 0.12$  & $0.47 \pm 0.0025$ \\
        \textit{KD-AdvReg} & $64.25 \pm 0.14$  & $56.73 \pm 0.14$  & $1.13 \pm 0.0008$\\
        \textit{KD-DMP} & $62.98 \pm 0.17$  & $69.08 \pm 0.19$ & $0.91 \pm 0.0003$ \\
        \textit{MIA-Pr} & $68.75 \pm 0.19$ & $62.84 \pm 0.24$ & $1.09 \pm 0.0014$ \\
        \textit{KL-Div} & $63.98 \pm 0.10$ & $55.10 \pm 0.32$ & $1.16 \pm 0.0050$  \\
        \midrule
        \textbf{\textit{BMIA-SafeCompress}} & $69.63 \pm 0.14$ & $52.04 \pm 0.21$ & $1.34 \pm 0.0048$\\
        \bottomrule
    \end{tabular}
    \vspace{-1em}
\end{table*}
\textbf{Time Consumption.} Table~\ref{time} reports the time consumption for each approach to get a compressed model (sparsity = 0.05) on CIFAR100. In general, almost all the approaches (except Pr-DP) can finish compression in 1--5 hours, and the time consumption of BMIA-SafeCompress is comparable to that of others. 
Considering that BMIA-SafeCompress can achieve the best performance-safety balance (from our previous experiment results), the time consumption of BMIA-SafeCompress is generally acceptable in practice. Note that we only need to run BMIA-SafeCompress once to get a compressed model, and the model can then be deployed in hundreds of thousands of devices repeatedly. 

Further, we provide a detailed analysis of training overhead below. In our framework, we only perform Stage 1 once, which requires little time (around $2s$ for VGG16) and thereby can be ignored. Stage 3 is simply about testing and also does not need much time. Hence, almost the whole training time is occupied by Stage 2 including sparse model training and attack mechanism simulation. Here, the time consumption of dynamic sparse update could be incorporated into sparse model training as the two steps are close: dynamic sparse update is triggered when the pre-defined iteration of sparse model training is over. We take VGG16 on CIFAR100 in Table~\ref{time} for example to illustrate the constitution of time consumption for each part. In our experiment, we find attack mechanism simulation consumes 2.8 hours (70\%). This consumption is more than 1.2 hours (30\%) of sparse model training. This result could be reasonable as attack mechanism simulation is reemployed for each update. It could be helpful to design new strategies of attack mechanism simulation to alleviate this issue. Since it is not our main focus, we will leave it as future work.

\textbf{Best-performing Selection Strategy in Safety Test.} In SafeCompress, we only keep the best-performing model in the safety testing due to time consumption. Given that we keep two models each time, the number of models sent back to ``Stage 2" will be double, \eg, 2, 4, 8, 16, and the time consumption for training will be exponentially growing. The cost is unbearable as the whole training process involves about $30$ times of updates, resulting in $2^{30}$ models left. Hence, we keep only the best-performing model. This choice will potentially result in the local optima. Nevertheless, considering the saved time consumption and that our approach performs the best among all solutions in various experiments, it is bearable for possibly being local optima. We also conduct an experiment by keeping no more than $4$ candidate models after the safety testing using VGG16 in CIFAR100 with sparsity set to 0.05. The whole time consumption is 15.36 hours, more than 4 times of our default setting (3.98 h). For the performance, this strategy produces $69.65\%$ for Task Acc and $51.72\%$ for MIA Acc. Compared to our default setting ($69.52\%$ Task Acc and $51.75\%$ MIA Acc), the improvement is marginal, empirically demonstrating that the best-performing selection strategy in SafeCompress is more reasonable in light of the performance and significantly saved time consumption.

\textbf{Flexibility with Other Training Tricks.}
It is worth noting that SafeCompress is also flexible to incorporate other training tricks not mentioned in previous experiments. To illustrate this flexibility, we try to incorporate five other widely-used training tricks in BMIA-SafeCompress, including \textit{dropout}~\cite{srivastava2014dropout}, \textit{data augmentation} (\textit{Aug}, including random cropping, resizing, and flipping), \textit{adversary regularization (AdvReg)}, \textit{sparse model initialization with the big model's parameter values (BigPara)}, and \textit{knowledge distillation (KD)} where the sparse model's prediction score approximates the big model's~\cite{KD_hinton2015distilling}. Table~\ref{flexsibility} shows the results of CIFAR100. 

Compared to our BMIA-SafeCompress implementation in main experiments, we find that most training tricks achieve similar TM-scores. The result indicates that SafeCompress is compatible with most training tricks. Then, in practice, given a specific dataset/task, we could enumerate different combinations of training tricks and find the best one in the SafeCompress framework. 

Another interesting observation is that \textit{data augmentation significantly increases MIA Acc} (\ie, reduces safety) for the final compressed model. The possible reason is that data augmentation allows training samples to be memorized more easily, as the variants of original samples (\eg, flipping and resizing) are also used for training. Notably, while not focusing on compressed sparse models, Yu~\etal also point out that data augmentation may lead to a significantly higher MIA risk~\cite{yu2021does}. Inspired by the work~\cite{yu2021does}, SafeCompress may also be extended to a useful framework to benchmark different training tricks' impacts on compressed models' safety. This task could be an interesting future direction.

\setlength{\tabcolsep}{0.35cm}\begin{table*}[h]
	\centering
	\caption{Defend against black-box MIA. Task Acc (performance), MIA Acc (safety) and TM-score$_{B}$ results on CIFAR10. The best results are marked in \underline{\textbf{bold}}.}
	\label{alexnet_0.05_0.1}
	\vspace{-1em}
	\begin{tabular}{l  c  c  c c c  c  c}
		\toprule
		\multirow{2}{*}{\textbf{Approach}}&
		\multicolumn{3}{c}{\textbf{Sparsity=0.05} }& &\multicolumn{3}{c}{\textbf{Sparsity=0.1}}\cr
		\cmidrule(lr){2-4} \cmidrule(lr){6-8}
		& \textit{Task Acc} & \textit{MIA Acc} & \textit{TM-score} &
		& \textit{Task Acc} & \textit{MIA Acc} & \textit{TM-score} \\
		\midrule
		\textit{AlexNet} (uncompressed) & 87.41\% & 58.34\% & 1.50 & & 87.41\% & 58.34\% & 1.50 \\
		\midrule
		\textit{Pr-DP} & 54.57\% & 68.35\% & 0.80 & & 62.87\% &66.49\% & 0.95  \\
		\textit{Pr-AdvReg} & 82.72\% &52.74\% & 1.57& & 86.19\%  & 54.83\% & 1.57  \\
		\textit{Pr-DMP} & 81.66\% & 55.63\% & 1.47 & & 83.73\% & 54.05\%& 1.55  \\
		\textit{KD-DP} & 72.09\% & 59.94\% & 1.20 & & 73.27\% & 60.42\% &  1.21 \\
		\textit{KD-AdvReg} & 80.47\% & 54.38\% & 1.48 & & $\underline{\textbf{86.84\%}}$ & 55.59\% & 1.56  \\
		\textit{KD-DMP} & 82.23\% & 61.63\% & 1.33 & & 86.26\% & 63.45\% & 1.36  \\
		\textit{MIA-Pr} & 82.26\% & 53.15\% & 1.55 & & 86.11\%& 56.12\% &  1.53 \\
		\textit{KL-Div} & 63.89\% &  $\underline{\textbf{52.00\%}}$ & 1.23 & & 72.19\% & 53.21\% &  1.36 \\
		\midrule
		\textbf{\textit{BMIA-SafeCompress}} & 83.94\% & 52.97\% &  $\underline{\textbf{1.59}}$ & & 85.31\% & $\underline{\textbf{52.37\%}}$  &$\underline{\textbf{1.63}}$  \\
		\textbf{\textit{+ Adversarial Training}} & $\underline{\textbf{84.20\%}}$ & 53.12\% & $\underline{\textbf{1.59}}$ & &85.38\% & 52.52\% & $\underline{\textbf{1.63}}$ \\
		\bottomrule
	\end{tabular}
\end{table*}

\setlength{\tabcolsep}{0.35cm}\begin{table*}
	\centering
	\caption{Defend against black-box MIA. Task Acc (performance), MIA Acc (safety) and TM-score$_{B}$ results on Tiny ImageNet. The best results are marked in \underline{\textbf{bold}}.}
	\vspace{-1em}
	\label{resnet_0.05_0.1}
	\begin{tabular}{l  c  c  c c c  c  c}
		\toprule
		\multirow{2}{*}{\textbf{Approach}}&
		\multicolumn{3}{c}{\textbf{Sparsity=0.05} }& &\multicolumn{3}{c}{\textbf{Sparsity=0.1}}\cr
		\cmidrule(lr){2-4} \cmidrule(lr){6-8}
		& \textit{Task Acc} & \textit{MIA Acc} & \textit{TM-score} &
		& \textit{Task Acc} & \textit{MIA Acc} & \textit{TM-score} \\
		\midrule
		\textit{ResNet18} (uncompressed) & 65.48\% & 69.73\% & 0.94 & &  65.48\% & 69.73\% & 0.94 \\
		\midrule
		\textit{Pr-DP} & 19.26\% & 71.07\% & 0.27 & & 24.56\% &74.17\% & 0.33  \\
		\textit{Pr-AdvReg} & 60.10\% & 57.62\% & 1.04 & & 61.16\%  & 63.93\% & 0.96  \\
		\textit{Pr-DMP} & 55.56\% & 60.03\% & 0.92 & & 59.61\% & 64.56\%& 0.92  \\
		\textit{KD-DP} & 17.10\% & 52.32\% & 0.33 & & 17.45\% & 52.44\% &  0.33 \\
		\textit{KD-AdvReg} & 52.34\% & 53.27\% & 0.98 & & 54.48\% & 53.60\% & 1.02  \\
		\textit{KD-DMP} & 53.71\% & 57.18\% & 0.94 & & 57.16\% & 55.65\% & 1.03  \\
		\textit{MIA-Pr} & 58.36\% & 57.92\% & 1.01 & & 60.91\%& 61.03\% &  1.00 \\
		\textit{KL-Div} & 48.22\% & 55.26\% & 0.87 & & 56.82\% & 57.19\% &  0.99 \\
		\midrule
		\textbf{\textit{BMIA-SafeCompress}} & $\underline{\textbf{63.81\%}}$ & 52.46\% & 1.22 & & $\underline{\textbf{65.15\%}}$ & 52.79\% & $\underline{\textbf{1.24}}$ \\
		 \textbf{\textit{+ Adversarial Training}} & 63.12\% &  $\underline{\textbf{51.38\%}}$  &$\underline{\textbf{1.23}}$ & &64.24\%  & $\underline{\textbf{52.01\%}}$ & $\underline{\textbf{1.24}}$ \\
		\bottomrule
	\end{tabular}
\end{table*}

\textbf{Scale to Big Model and Large Dataset.} In SafeCompress, we use CIFAR10, CIFAR100, and Tiny ImageNet in our experiment as they are widely used in membership inference attacks in previous work~\cite{liu2021encodermi, shokri2017membership, salem2018ml, kaya2021does}. To demonstrate the scalability, we scale model size by using ResNet50~\cite{resnet_he2016deep} and train on a large dataset ImageNet~\cite{deng2009imagenet}. We set sparsity to 0.1 following previous work~\cite{DST_InTime} with a black-box setting. In our experiment, BMIA-SafeCompress obtains $73.80\%$ Task Acc and $56.67\%$ MIA Acc, resulting in $1.3$ of TM-score which is superior to ResNet50 (uncompressed) under the same 100 training epochs whose TM-score is $1.16$ ($76.13\%$ Task Acc and $65.23\%$ MIA Acc). The result indicates that our approach is also scalable.

\textbf{Robustness on Metric Variants over Baselines.} We can achieve different requirements for safety and performance in our approach by adjusting the coefficient $\lambda$ (\eg, 0.8, 0.9, 1.0, 1.1, and 1.2) of the TM-score.  To show whether our approach is robust to this situation over the other eight baselines, we illustrate the results of CIFAR100 in Table~\ref{lambda_tm-score}. It is observed that BMIA-SafeCompress performs the best over other baselines among all the settings. This result indicates that BMIA-SafeCompress is capable of meeting different degrees of safety and performance balance over baselines. This result also implies that our SafeCompress may naturally have the property of robustness against metric alteration to some extent.

\textbf{Statistical Analysis with Multiple Runs.} Due to the stochastic nature of optimization algorithms and DNN training procedures, we conduct multiple runs in a case to present a statistical analysis. We use BMIA-Safecompress using VGG16 on CIFAR100 with sparsity set to 0.05. We run each approach three times with random seeds and report the results in Tab~\ref{vgg_mutiple_run}. One can see that our approach is statistically superior to others. To further support our conclusion, we perform a statistical test using the student's t test (considering the sample numbers) and compare the P value with that of the best-performing baseline on each metric. For Task Acc, we compare BMIA-SafeCompress with MIA-Pr, and the P value is 0.007, smaller than the default alpha level (0.01). For MIA Acc, we compare BMIA-SafeCompress with KL-Div, and the P value is 0.0004 ($0.0004 < 0.01$). As for TM-Score, we compare BMIA-SafeCompress with KL-Div. The P value ($1.2e-5$) is also smaller than 0.01. These results show our conclusion is significant.

\setlength{\tabcolsep}{0.25cm}\begin{table}[h]
	\caption{Defend against black-box MIA. Task Acc, MIA Acc, and TM-score$_{B}$ on Yelp-5.}
	\label{bert_0.5}
	\begin{tabular}{l  c  c  c}
		\toprule
		\multirow{2}{*}{\textbf{Approach}}&
		\multicolumn{3}{c}{\textbf{Sparsity=0.5} }\cr
		\cmidrule(lr){2-4}
		& \textit{Task Acc} & \textit{MIA Acc} & \textit{TM-score}
		\\
		\midrule
		\textit{BERT} (Uncompressed) & 62.21%
		\% & 71.15\% & 0.87 \\
		\midrule
		\textit{Pr-DP} & 60.77\% & 64.79\% & 0.94  \\
		\textit{Pr-AdvReg} & 61.17\% & 73.34\% & 0.83\\
		\textit{Pr-DMP} & 61.32\% & 66.41\% & 0.92  \\
		\textit{KD-DP} & 58.21\% & 65.36\% & 0.89  \\
		\textit{KD-AdvReg} & 61.38\% & 67.40\% & 0.91 \\
		\textit{KD-DMP} & 59.69\% & 70.25\% & 0.85 \\
		\textit{MIA-Pr} & 61.80\% & 73.85\% & 0.84 \\
		\textit{KL-Div} &  60.03\% & $\underline{\textbf{64.24\%}}$ &  0.93\\
		\midrule
		\textit{\textbf{BMIA-SafeCompress}} & 61.66\% & 64.37\% &  0.96 \\
		\textit{\textbf{+ Adversarial Training}} & $\underline{\textbf{62.63\%}}$  &  64.90\% &$\underline{\textbf{0.97}}$ \\
		\bottomrule
	\end{tabular}
\end{table}

\subsubsection{Results on CV Datasets}
\textbf{CIFAR10 (AlexNet).} 
The results on CIFAR10 are presented in Table~\ref{alexnet_0.05_0.1}. It can be seen that BMIA-SafeCompress obtains almost the best performance in Task Acc and maintains a pretty strong defensive ability when sparsity is 0.05. Thanks to the excellent effectiveness in both aspects, BMIA-SafeCompress produces the highest TM-score, showing its outstanding ability to make the performance-safety trade-off. When the sparsity is set to 0.1, our approach produces 85.31\% for Task ACC, slightly inferior to the best result (86.84\%) produced by KD-AdvReg. However, BMIA-SafeCompress decreases MIA Acc to 52.37\%, much lower than KD-AdvReg (55.59\%), leading to the highest TM-score again. Finally, leveraging Adversarial Training leads to further improvement in Task Acc at the cost of a slight increase in MIA Acc.

\setlength{\tabcolsep}{0.35cm}\begin{table*}[h]
	\centering
	\caption{Defend against white-box MIA. Task Acc (performance), MIA Acc (safety) and TM-score$_{W}$ results on CIFAR10. The best results are marked in \underline{\textbf{bold}}.}
	\label{alexnet_0.05_0.1_w}
	\begin{tabular}{l  c  c  c c c  c  c}
		\toprule
		\multirow{2}{*}{\textbf{Approach}}&
		\multicolumn{3}{c}{\textbf{Sparsity=0.05} }& &\multicolumn{3}{c}{\textbf{Sparsity=0.1}}\cr
		\cmidrule(lr){2-4} \cmidrule(lr){6-8}
		& \textit{Task Acc} & \textit{MIA Acc} & \textit{TM-score} &
		& \textit{Task Acc} & \textit{MIA Acc} & \textit{TM-score} \\
		\midrule
		\textit{AlexNet} (uncompressed) & 87.41\% & 59.13\% & 1.48 & & 87.41\% & 59.13\% & 1.48 \\
		\midrule
		\textit{Pr-DP} & 54.57\% & 52.83\% & 1.03  & & 62.87\% & 54.23\% & 1.16 \\
		\textit{Pr-AdvReg} & 82.72\% & 54.84\% & 1.51 & & 86.19\%  & 57.82\% & 1.49 \\
		\textit{Pr-DMP} & 81.66\% & 53.69\% & 1.52 & & 83.73\% & 54.84\%&  1.53 \\
		\textit{KD-DP} & 72.09\% & $\underline{\textbf{50.95\%}}$ & 1.41 & & 73.27\% & $\underline{\textbf{50.88\%}}$  &  1.44 \\
		\textit{KD-AdvReg} & 80.47\% & 54.91\% & 1.47 & & $\underline{\textbf{86.84\%}}$ & 56.95\% &  1.52 \\
		\textit{KD-DMP} & 82.23\% & 54.27\% & 1.52 & & 86.26\% & 54.09\% & 1.59  \\
		\textit{MIA-Pr} & 82.26\% & 55.48\% & 1.48 & & 86.11\%& 58.14\% & 1.48 \\
		\textit{KL-Div} & 63.89\% & 52.32\% & 1.22 & & 72.19\%& 53.32\% & 1.35 \\
		\midrule
		\textbf{\textit{WMIA-SafeCompress}} & 81.33\%  & 53.03\% & 1.53 & & 83.08\% & 52.38\%  & $\underline{\textbf{1.59}}$  \\
		\textbf{\textit{+ Adversarial Training}} & $\underline{\textbf{83.38\%}}$ & 53.32\% & $\underline{\textbf{1.56}}$ & &84.70\% & 53.34\% & $\underline{\textbf{1.59}}$ \\
		\bottomrule
	\end{tabular}
\end{table*}
\setlength{\tabcolsep}{0.35cm}\begin{table*}[h]
	\centering
	\caption{Defend against white-box MIA. Task Acc (performance), MIA Acc (safety) and TM-score$_{W}$ results on Tiny ImageNet. The best results are marked in \underline{\textbf{bold}}.}
	\label{resnet_0.05_0.1_w}
	\begin{tabular}{l  c  c  c c c  c  c}
		\toprule
		\multirow{2}{*}{\textbf{Approach}}&
		\multicolumn{3}{c}{\textbf{Sparsity=0.05} }& &\multicolumn{3}{c}{\textbf{Sparsity=0.1}}\cr
		\cmidrule(lr){2-4} \cmidrule(lr){6-8}
		& \textit{Task Acc} & \textit{MIA Acc} & \textit{TM-score} &
		& \textit{Task Acc} & \textit{MIA Acc} & \textit{TM-score} \\
		\midrule
		\textit{ResNet18} (uncompressed) & 65.48\% & 80.15\% & 0.82 & &  65.48\% & 80.15\% & 0.82 \\
		\midrule
		\textit{Pr-DP} & 19.26\% & 60.88\% & 0.32 & & 24.56\% & 66.98\% & 0.37  \\
		\textit{Pr-AdvReg} & 60.10\% & 63.77\% & 0.94 & & 61.16\%  & 70.37\% & 0.87 \\
		\textit{Pr-DMP} & 55.56\% & 57.72\% & 0.96 & & 59.61\% & 58.72\%& 1.02 \\
		\textit{KD-DP} & 17.10\% & $\underline{\textbf{51.09\%}}$ & 0.33 & & 17.45\% & $\underline{\textbf{51.17\%}}$&  0.34 \\
		\textit{KD-AdvReg} & 52.34\% & 58.08\% & 0.90 & & 54.48\% & 54.81\% & 0.99  \\
		\textit{KD-DMP} & 53.71\% & 55.65\% & 0.97 & & 57.16\% & 54.45\% & 1.05  \\
		\textit{MIA-Pr} & 58.36\% & 58.66\% & 0.99 & & 60.91\% & 69.47\% & 0.88 \\
		\textit{KL-Div} & 48.22\% & 54.78\% & 0.88 & & 56.82\% & 54.96\% & 1.03 \\
		\midrule
		\textbf{\textit{WMIA-SafeCompress}} & $\underline{\textbf{60.86\%}}$ & 53.09\% & $\underline{\textbf{1.15}}$ & & $\underline{\textbf{61.57\%}}$ & 54.23\% & 1.14 \\
		\textbf{\textit{+ Adversarial Training}} & 59.50\% &  51.85\% &$\underline{\textbf{1.15}}$ & &60.81\%  & 51.72\% & $\underline{\textbf{1.18}}$ \\
		\bottomrule
	\end{tabular}
\end{table*}

\setlength{\tabcolsep}{0.2cm}\begin{table}[h]
	\centering
	\caption{Defend against white-box MIA. Task Acc, MIA Acc, and TM-score$_{W}$ on AG News.}
	\label{roberta_0.5_w}
	\begin{tabular}{l  c  c  c}
		\toprule
		\multirow{2}{*}{\textbf{Approach}}&
		\multicolumn{3}{c}{\textbf{Sparsity=0.5} }\cr
		\cmidrule(lr){2-4}
		& \textit{Task Acc} & \textit{MIA Acc} & \textit{TM-score}
		\\
		\midrule
		\textit{RoBERTa} (Uncompressed) & 89.20\% & 57.43\% &  1.55\\
		\midrule
		\textit{Pr-DP} & 87.28\% & 55.96\% &  1.56\\
		\textit{Pr-AdvReg} & 87.24\% & 56.12\% &  1.55\\
		\textit{Pr-DMP} & 86.38\% & 55.28\% & 1.56\\
		\textit{KD-DP} & 82.64\% & $\underline{\textbf{53.02\%}}$ &  1.56\\
		\textit{KD-AdvReg} & $\underline{\textbf{88.10\%}}$ & 56.33\% & 1.56\\
		\textit{KD-DMP} & 87.24\% & 55.95\% &   1.56 \\
		\textit{MIA-Pr} & 87.91\% &  57.25\%& 1.54\\
		\textit{KL-Div} & 83.26\% &  54.18\%& 1.54\\
		\midrule
		\textit{\textbf{WMIA-SafeCompress}} & 87.31\% & 55.75\% &   1.57\\
		 \textit{\textbf{+ Adversarial Training}} & 87.63\% &  55.06\%  &$\underline{\textbf{1.59}}$ \\
		\bottomrule
	\end{tabular}
\end{table}

\textbf{Tiny ImageNet (ResNet18).}  
We present the results of Tiny ImageNet in Table~\ref{resnet_0.05_0.1}. Excitingly, BMIA-SafeCompress produces the best accuracy for Task Acc in both sparsity 0.05 and 0.1, outperforming all the baselines by a large margin. In addition, BMIA-SafeCompress obtains competitive MIA Acc in both sparsity settings compared to the best defense effects of baselines. Specifically, BMIA-SafeCompress achieves 52.46\% (sparsity 0.05) and 52.79\% (sparsity 0.1) in MIA Acc; The best MIA Acc results among baselines are 52.32\% (sparsity 0.05) and 52.44\% (sparsity 0.1). The gaps are tiny (0.14\% and 0.35\%). Finally, BMIA-SafeCompress (+ Adversarial Training) achieves the best TM-score in both sparsity settings. 

\subsubsection{Results on NLP Dataset}
\textbf{Yelp-5 (BERT).} The results are reported in Table~\ref{bert_0.5}. It can be seen that BMIA-SafeCompress produces a competitive Task Acc. At the same time, it decreases MIA Acc by 6.78\% compared to the uncompressed model, outperforming all the baselines. BMIA-SafeCompress produces 0.96 for the TM-score, indicating its ability to balance task performance and safety. Besides, when attaching Adversarial Training to BMIA-SafeCompress, we further improve the Task Acc, leading to the highest TM-score.

\setlength{\tabcolsep}{0.2cm}\begin{table*}[h]
	\centering
	\caption{Defend against black-box and white-box MIAs. Task Acc (performance), MIA Acc (safety) and TM-score$_{M}$ results on CIFAR10. The best results are marked in \underline{\textbf{bold}}.}
	\label{alexnet_0.05_0.1_m}
	\begin{tabular}{l  c  c  c c  c  c c  c  c}
		\toprule
		\multirow{2}{*}{\textbf{Approach}}&
		\multicolumn{4}{c}{\textbf{Sparsity=0.05} }& &\multicolumn{4}{c}{\textbf{Sparsity=0.1}}\cr
		\cmidrule(lr){2-5} \cmidrule(lr){7-10}
		& \textit{Task Acc} & \textit{MIA Acc$_{B}$} & \textit{MIA Acc$_{W}$} & \textit{TM-score} &
		& \textit{Task Acc} & \textit{MIA Acc$_{B}$} & \textit{MIA Acc$_{W}$} & \textit{TM-score} \\
		\midrule
		\textit{AlexNet} (uncompressed) & 87.41\% & 58.34\% & 59.13\% & 1.49 & & 87.41\% & 58.34\% & 59.13\% & 1.49 \\
		\midrule
		\textit{Pr-DP} & 54.57\% & 68.35\% & 52.83\% & 0.92 & & 62.87\% & 66.49\% & 54.23\% & 1.05 \\
		\textit{Pr-AdvReg} & 82.72\% & 52.74\% & 54.84\% & 1.53 & & 86.19\%  & 54.83\% & 57.82\% &  1.53 \\
		\textit{Pr-DMP} & 81.66\% & 55.63\% & 53.69\% & 1.49 & & 83.73\% & 54.05\%& 54.84\% & 1.54 \\
		\textit{KD-DP} & 72.09\% & 59.94\% & $\underline{\textbf{50.95\%}}$ & 1.31 & & 73.27\% & 60.42\% & $\underline{\textbf{50.88\%}}$ & 1.33 \\
		\textit{KD-AdvReg} & 80.47\% & 54.38\% & 54.91\% & 1.52 & & $\underline{\textbf{86.84\%}}$ & 55.59\% & 56.95\% & 1.54 \\
		\textit{KD-DMP} & 82.23\% & 61.63\% & 54.27\% & 1.42 & & 86.26\% & 63.45\% & 54.09\% & 1.48 \\
		\textit{MIA-Pr} & 82.26\% & 53.15\% & 55.48\% & 1.52 & & 86.11\%& 56.12\% & 58.14\% & 1.51 \\
		\textit{KL-Div} & 63.89\% & 52.00\% & 52.32\% & 1.22  & & 72.19\% & 53.21\% & 53.32\% & 1.36 \\
		\midrule
		\textbf{\textit{MMIA-SafeCompress}} & 81.64\% & 51.79\% & 54.98\% & 1.53 & & 83.98\% & $\underline{\textbf{52.81\%}}$ & 55.86\% & 1.55 \\
		\textbf{\textit{+ Adversarial Training}} & $\underline{\textbf{83.29\%}}$ & $\underline{\textbf{50.28\%}}$  & 54.70\% & $\underline{\textbf{1.59}}$ &  & 85.12\% & 54.21\%  & 54.94\% & $\underline{\textbf{1.56}}$ \\
		\bottomrule
	\end{tabular}
\end{table*}
\setlength{\tabcolsep}{0.2cm}\begin{table*}[h]
	\centering
	\caption{Defend against black-box and white-box MIAs. Task Acc (performance), MIA Acc (safety) and TM-score$_{M}$ results on Tiny ImageNet. The best results are marked in \underline{\textbf{bold}}.}
	\vspace{-1em}
	\label{resnet_0.05_0.1_m}
	\begin{tabular}{l  c  c  c c  c  c c  c  c}
		\toprule
		\multirow{2}{*}{\textbf{Approach}}&
		\multicolumn{4}{c}{\textbf{Sparsity=0.05} }& &\multicolumn{4}{c}{\textbf{Sparsity=0.1}}\cr
		\cmidrule(lr){2-5} \cmidrule(lr){7-10}
		& \textit{Task Acc} & \textit{MIA Acc$_{B}$} & \textit{MIA Acc$_{W}$} & \textit{TM-score} &
		& \textit{Task Acc} & \textit{MIA Acc$_{B}$} & \textit{MIA Acc$_{W}$} & \textit{TM-score} \\
		\midrule
		\textit{ResNet18} (uncompressed) & 65.48\% & 69.73\% & 80.15\% & 0.88 & & 65.48\% & 69.73\% & 80.15\% & 0.88 \\
		\midrule
		\textit{Pr-DP} & 19.26\% & 71.07\% & 60.88\% & 0.29 & & 24.56\% & 74.17\% & 66.98\% & 0.35 \\
		\textit{Pr-AdvReg} & 60.10\% & 57.62\% & 63.77\% & 0.99 & & 61.16\%  & 63.93\% &  70.37\% & 0.91 \\
		\textit{Pr-DMP} & 55.56\% & 60.30\% & 57.72\% & 0.94 & & 59.61\% & 64.56\%& 58.72\% & 0.97 \\
		\textit{KD-DP} & 17.10\% & 52.32\% & $\underline{\textbf{51.09\%}}$ & 0.33 & & 17.45\% & $\underline{\textbf{52.44\%}}$ &  $\underline{\textbf{51.17\%}}$ & 0.34  \\
		\textit{KD-AdvReg} & 52.34\% & 53.27\% & 58.08\% & 0.94 & & 54.48\% & 53.60\% & 54.81\% & 1.01 \\
		\textit{KD-DMP} & 53.71\% & 57.18\% & 55.65\% & 0.95 & & 57.16\% & 55.65\% & 54.45\% & 1.04\\
		\textit{MIA-Pr} & 58.36\% & 57.92\% & 58.66\% & 1.00 & & 60.91\%& 61.03\% & 69.47\% & 0.94  \\
		\textit{KL-Div} & 48.22\% & 55.26\% & 54.78\% & 0.88 & & 56.82\% & 57.19\% & 54.96\% & 1.01 \\
		\midrule
		\textbf{\textit{MMIA-SafeCompress}} & $\underline{\textbf{62.23\%}}$ & 50.59\% & 54.13\% & $\underline{\textbf{1.19}}$ & & $\underline{\textbf{63.53\%}}$ & 53.09\%  & 56.34\% & $\underline{\textbf{1.16}}$ \\
		\textbf{\textit{+ Adversarial Training}} & 61.76\% & $\underline{\textbf{50.28\%}}$ & 53.55\% & $\underline{\textbf{1.19}}$ &  & 62.51\% & 52.77\%  & 55.94\% & 1.15 \\
		\bottomrule
	\end{tabular}
\end{table*}

\subsection{Experimental Results on WMIA-SafeCompress}
\subsubsection{Results on CV Datasets}
\textbf{CIFAR10 (AlexNet).} We provide results as shown in Table~\ref{alexnet_0.05_0.1_w} to further show the effectiveness. It can be seen that WMIA-SafeCompress obtains a very competitive performance in Task Acc and maintains a pretty strong defensive ability when sparsity is 0.05. As a result, WMIA-SafeCompress produces the highest TM-score, showing its outstanding ability to make the performance-safety trade-off. When the sparsity is set to 0.1, our approach produces 83.08\% for Task ACC, inferior to the best result (86.84\%) produced by KD-AdvReg. However, WMIA-SafeCompress decreases MIA Acc to 52.38\%, much lower than KD-AdvReg (56.95\%), leading to the highest TM-score again (the same as KD-DMP). Finally, incorporating Adversarial Training with WMIA-SafeCompress leads to a further improvement in Task Acc at the cost of a relatively slight increase of MIA Acc, especially for sparsity 0.05.

\textbf{Tiny ImageNet (ResNet18).} The results on Tiny ImageNet here as shown in Table~\ref{resnet_0.05_0.1_w}.  Excitingly, WMIA-SafeCompress produces the best accuracy for Task Acc in both sparsity 0.05 and 0.1, outperforming all the baselines by a large margin. Additionally, WMIA-SafeCompress obtains relatively low MIA Acc in both sparsity settings compared to the defense effects of baselines. Specifically, the best MIA Acc results among baselines are 51.09\% (sparsity 0.05) and 51.17\% (sparsity 0.1) that are both produced by KD-DP costing at intolerable performance drop; The worst MIA Acc results among baselines are 63.77\% (sparsity 0.05) and 70.37\% (sparsity 0.1) that are both produced by Pr-AdvReg; In contrast, WMIA-SafeCompress performs slightly well, achieving 53.09\% (sparsity 0.05) and 54.23\% (sparsity 0.1) in MIA Acc and obtaining the best TM-score in both sparsity settings. Finally, equipping with Adversarial Training greatly decreases privacy risk, especially for sparsity 0.1 where the TM-score rises to 1.18, outperforming other baselines by a large margin. 

\subsubsection{Results on NLP Dataset}

\textbf{AG News (RoBERTa).} To further validate the effectiveness and generalization of WMIA-SafeCompress, we conduct experiments on AG News. As indicated in Table~\ref{roberta_0.5_w}, WMIA-SafeCompress obtains a competitive performance in Task Acc achieving 87.31\%, slightly inferior
to the highest 88.10\% produced by KD-AdvReg. However, our approach
decreases MIA Acc to 55.75\%, lower than KD-AdvReg
(56.33\%). When comparing with KD-DP, we observe that WMIA-SafeCompress is a little bit inferior to KD-DP in defense, but outperforms KL-Div by a large margin in Task Acc. As a result, WMIA-SafeCompress achieves the highest TM-score, showing its great trade-off capability. Finally, leveraging Adversarial Training leads to a further enhancement in all the metrics.

\subsection{Experimental Results on MMIA-SafeCompress}
\subsubsection{Results on CV Datasets}
\textbf{CIFAR10 (AlexNet).} We present results on CIFAR10 in Table~\ref{alexnet_0.05_0.1_m}. It is observed that MMIA-SafeCompress achieves competitive classification performance among all baselines in both sparsity. It also decreases MIA Acc$_B$ and MIA Acc$_W$ compared to the uncompressed model, greatly improving the defense capability of the compressed model. Further, we can see that MMIA-SafeCompress produces the highest TM-score, demonstrating its outstanding capability to make the performance-safety trade-off. Besides, equipping MMIA-SafeCompress with Adversarial Training leads to great improvement in Task Acc and generally strengthens model defense ability. Consequently, these enhancements contribute to further boosting in TM-score.

\textbf{Tiny ImageNet (ResNet18).} The results on Tiny ImageNet are presented in Table~\ref{resnet_0.05_0.1_m}. It is easily seen that MMIA-SafeCompress produces 62.23\% (sparsity=0.05) and 63.53\% (sparsity=0.1) for Task Acc, outperforming all baselines by a large margin. In addition, MMIA-SafeCompress maintains considerable defense capability. For example, it holds the lowest MIA Acc$_B$ and the second-lowest MIA Acc$_W$ when the sparsity is set to 0.05. Thus, MMIA-SafeCompress achieves the highest TM-score in both sparsity and even outperforms the other eight approaches by a large margin. When combined with Adversarial Training, MMIA-SafeCompress further enhances defense ability via costing at a slight performance drop and brings little bad influence to the final TM-score. 

\subsubsection{Results on NLP Datasets}
\textbf{AG News (RoBERTa).} We further conduct experiments on AG News to validate the effectiveness and generalization of MMIA-SafeCompress. The results are reported in Table~\ref{roberta_0.5_m}. It can be seen that MMIA-SafeCompress achieves a competitive performance in Task Acc among all the approaches. Moreover, MMIA-SafeCompress maintains
considerable defense capability. For example, it holds the second-lowest MIA Acc$_B$ and the third-lowest MIA Acc$_W$. Consequently, these advantages lead MMIA-SafeCompress to
the highest TM-score. Further, combining MMIA-SafeCompress with
Adversarial Training improves the task performance and 
defense ability against the black-box attack while causing a slightly increased risk for white-box attack. Finally, such alterations bring little bad influence on TM-score.

\setlength{\tabcolsep}{0.02cm}\begin{table}[h]
	\centering
	\caption{Defend against black-box and white-box MIAs. Task Acc, MIA Acc$_{B}$, MIA Acc$_{W}$, and TM-score$_{M}$ on AG News.}
	\label{roberta_0.5_m}
	\begin{tabular}{l  c  c c  c}
		\toprule
		\multirow{2}{*}{\textbf{Approach}}&
		\multicolumn{3}{c}{\textbf{Sparsity=0.5} }\cr
		\cmidrule(lr){2-5}
		& \textit{Task Acc} & \textit{MIA Acc$_{B}$} & \textit{MIA Acc$_{W}$} & \textit{TM-score}
		\\
		\midrule
		\textit{RoBERTa} (Uncompressed) & 89.20\% & 57.59\% &  57.43\% & 1.55\\
		\midrule
		\textit{Pr-DP} & 87.28\% & 56.07\% & 55.96\% &  1.56\\
		\textit{Pr-AdvReg} & 87.24\% & 57.58\% & 56.12\% & 1.53\\
		\textit{Pr-DMP} & 86.38\% & 56.31\% & 55.28\% &  1.55\\
		\textit{KD-DP} & 82.64\% & 56.13\% & $\underline{\textbf{53.02\%}}$  &  1.52\\
		\textit{KD-AdvReg} & $\underline{\textbf{88.10}}$\% & 56.56\% & 56.33\% &  1.56\\
		\textit{KD-DMP} & 87.24\% & 56.64\% & 55.95\% & 1.55\\
		\textit{MIA-Pr} & 87.91\% & 57.04\% & 57.25\% & 1.54\\
		\textit{Kl-Div} & 83.26\% & $\underline{\textbf{54.09\%}}$ & 54.18\% & 1.54\\
		\midrule
		\textit{\textbf{MMIA-SafeCompress}} & 87.04\% & 55.92\% & 55.04\% & $\underline{\textbf{1.57}}$ \\
		\textbf{\textit{+ Adversarial Training}} & 87.31\% & 55.45\% & 55.65\% & $\underline{\textbf{1.57}}$ \\
		\bottomrule
	\end{tabular}
\end{table}

\section{}
\subsection{Discussion about trade-off between performance, safety, compressed models, and time costs}

First, the size to which a big model should be compressed is determined by the scenario or the requirements. For example, to deploy AI software on smartphones, when one hopes the deployed size is 10\% of an original uncompressed one, the sparsity is 0.1. Secondly, to balance the trade-off between performance and safety, we can simply adjust the coefficient $\lambda$ of the TM-score in Eq (6). For example, a user may think safety is more important than performance, and therefore she can decrease $\lambda$ to a smaller value; Otherwise, she can increase $\lambda$ to a bigger value. In most experiments, we set $\lambda=1$ by default as we consider performance and safety to be equally important. We also conduct experiments with different $\lambda$ presented in Figure 5 in our article. As for time costs, it is related to the training times of SafeCompress. Theoretically, we can stop at any time. In general, the more the training time is (if we do not consider overfitting), the better the performance is. Hence, we could predefine a number of training iterations to balance the time consumption and performance.


%

\ifCLASSOPTIONcompsoc


\bibliographystyle{IEEEtran}
\bibliography{bibfile}
%

%

\begin{IEEEbiography}[{\includegraphics[width=1in,height=1.25in,clip,keepaspectratio]{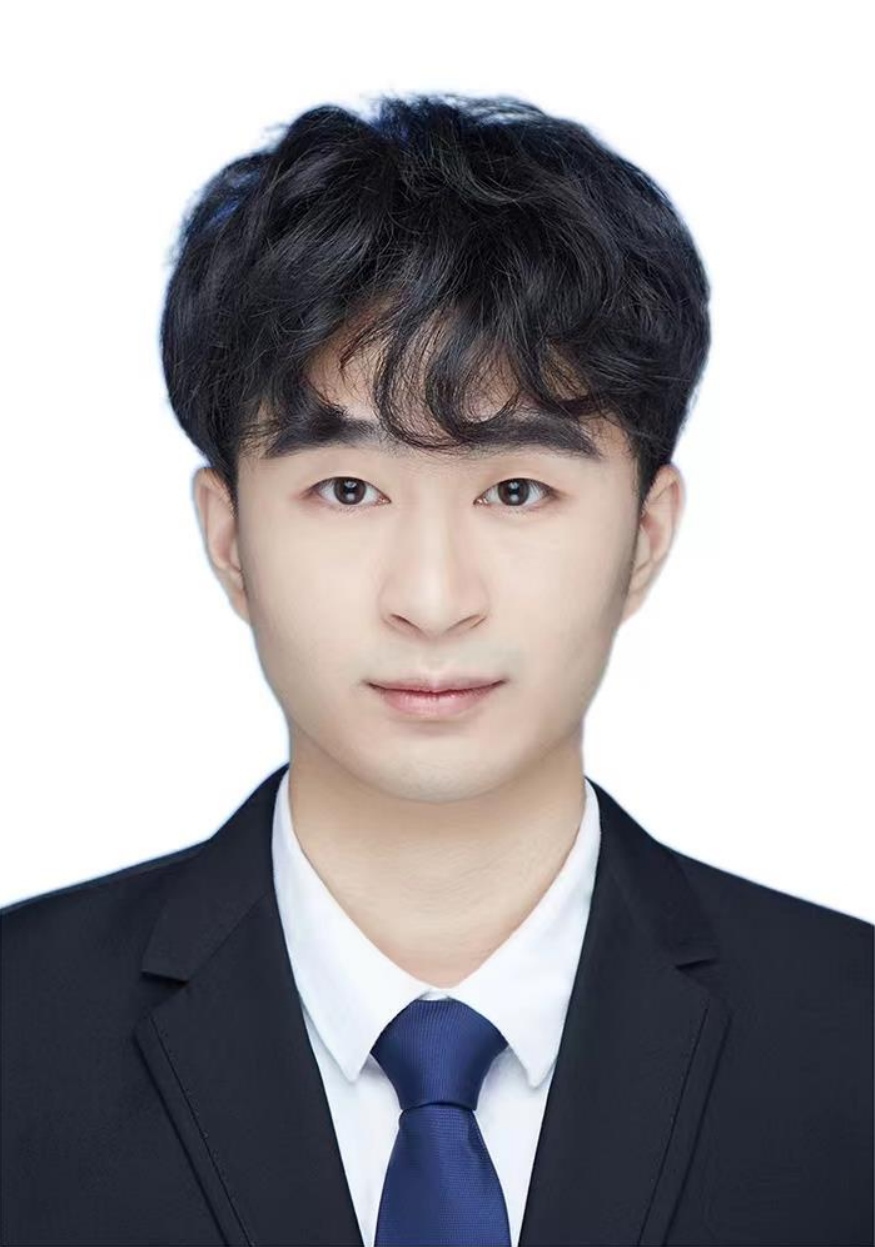}}]{Jie Zhu}
is currently pursuing the Ph.D. degree in the School of Computer Science, Peking University, China.
\end{IEEEbiography}
\vspace{-0.5cm}
\begin{IEEEbiography}[{\includegraphics[width=1in,height=1.25in,clip,keepaspectratio]{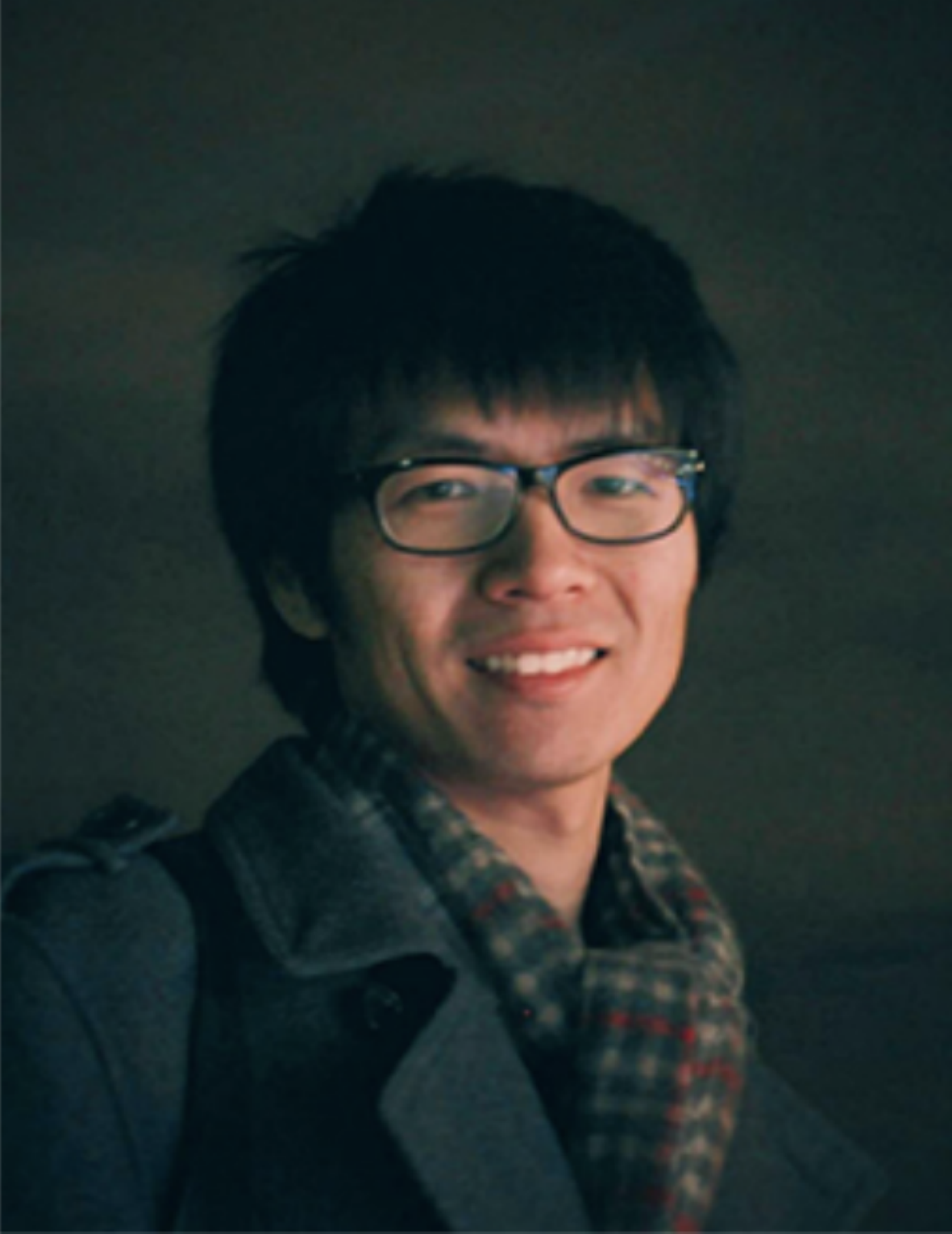}}]{Leye Wang} 
received the Ph.D. degree in computer science from TELECOM SudParis and University Paris 6, France, in 2016. He is currently an
assistant professor with the Key Lab of High Confidence Software Technologies, Peking University, MOE, and the School of Computer Science,
Peking University, China. He was a postdoctoral
researcher with the Hong Kong University of Science and Technology. His research interests
include ubiquitous computing, mobile crowdsensing, and urban computing.
\end{IEEEbiography}
\vspace{-0.5cm}

\begin{IEEEbiography}[{\includegraphics[width=1in,height=1.25in,clip,keepaspectratio]{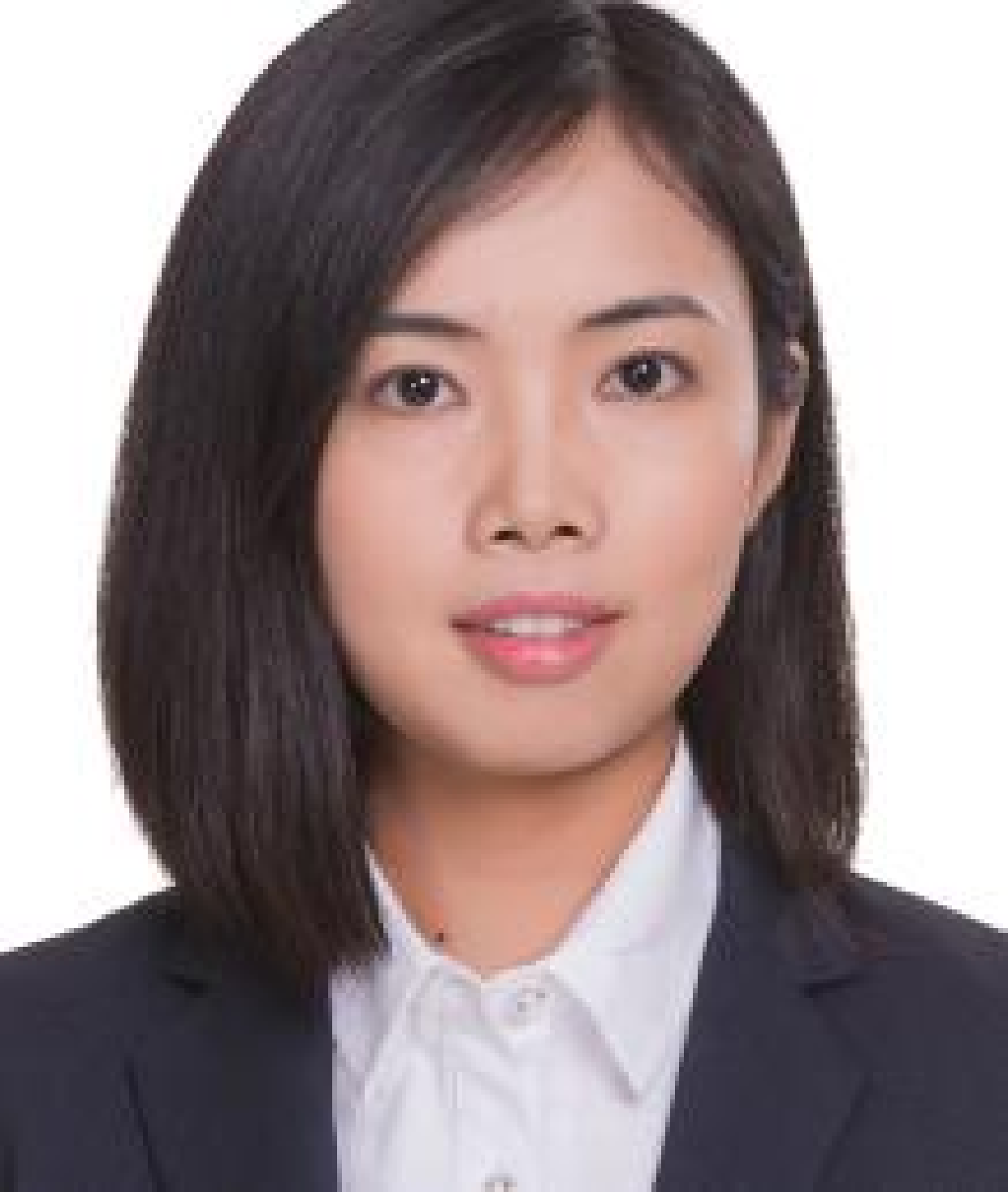}}]{Xiao Han} 
received the Ph.D. degree in computer
science from Pierre and Marie Curie University and
the Institut Mines-TELECOM/TELECOM SudParis
in 2015. She is currently a Full Professor with
the Shanghai University of Finance and Economics,
China. Her research interests include social network
analysis, fintech, and privacy protection.
\end{IEEEbiography}

\vspace{-0.5cm}

\begin{IEEEbiography}[{\includegraphics[width=1in,height=1.25in,clip,keepaspectratio]{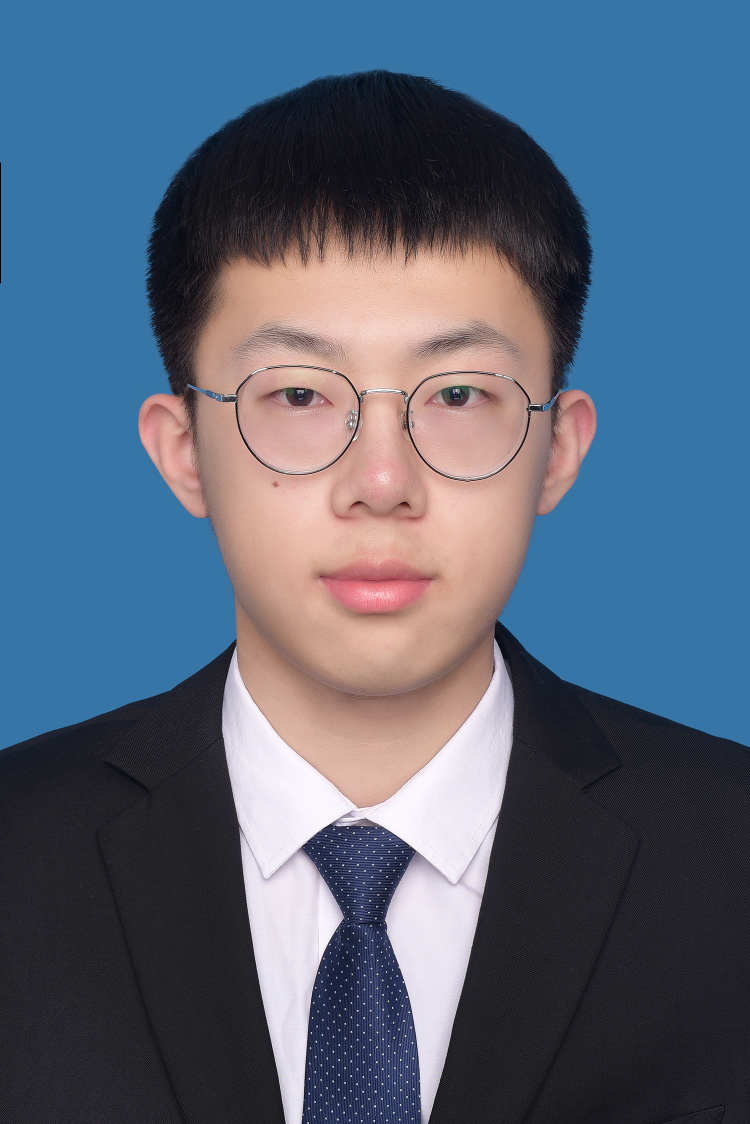}}]{Anmin Liu}
is currently pursuing the Ph.D. degree in the School of Computer Science, Peking University, China.
\end{IEEEbiography}

\vspace{-0.5cm}

\begin{IEEEbiography}[{\includegraphics[width=1in,height=1.25in,clip,keepaspectratio]{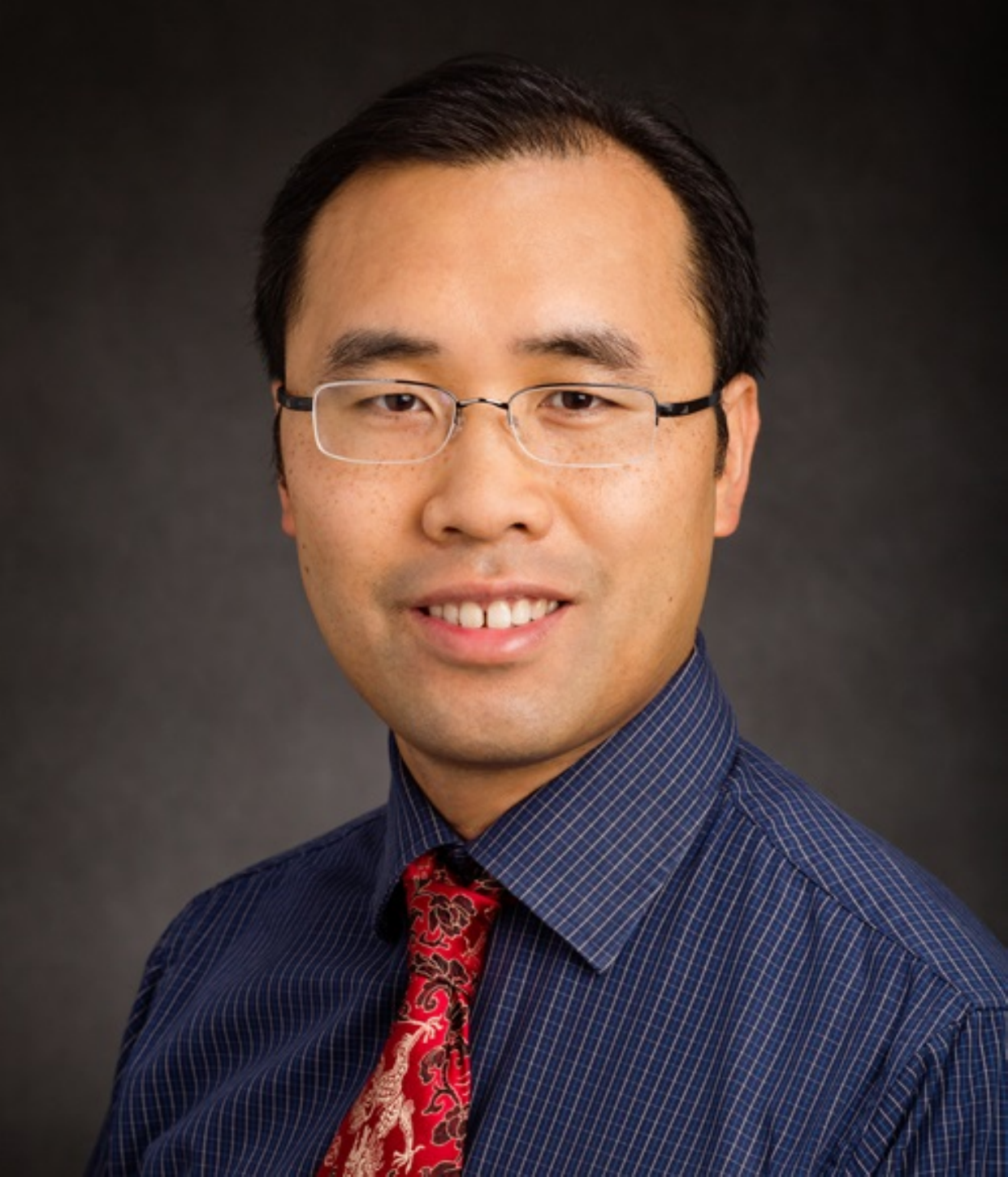}}]
{Tao Xie} received the BS degree in computer science from Fudan  University, Shanghai, China, the MS degree in computer science from Peking University, Beijing, China, and the PhD degree in computer science from the University of Washington at Seattle, WA. He is a Peking University Chair Professor, where he specializes in software engineering, system software, software security, and trustworthy AI. He is a Foreign Member of Academia Europaea, and a Fellow of ACM, IEEE, and AAAS. 
\end{IEEEbiography}



\end{document}